\pdfoutput=1
\documentclass{article}

\usepackage[colorlinks=true,urlcolor=blue,anchorcolor=blue,citecolor=blue,filecolor=blue,linkcolor=blue,menucolor=blue,linktocpage=true]{hyperref}
\usepackage{url}
\usepackage{amsfonts}       
\usepackage{nicefrac}       
\usepackage{microtype}      
\usepackage{graphicx}       
\usepackage{subcaption}
\usepackage{layouts}
\usepackage{xcolor}
\usepackage{amsmath}
\usepackage{amssymb}
\usepackage{amsthm}
\usepackage{booktabs}       
\usepackage{enumitem}       
\usepackage{wrapfig}

\usepackage{amsmath}

\usepackage{authblk}
\usepackage{fullpage}

\usepackage{float}

\usepackage{cuted}
\usepackage{breqn}

\usepackage[round]{natbib}

\usepackage{listings}         

\newcommand{\removed}[1]{}

\title{\vspace{0.0cm}\bf{What does a deep neural network confidently perceive? The effective dimension of high certainty class manifolds and their low confidence boundaries}}

\author[1]{Stanislav Fort\footnote{Now at Anthropic. Work done while at Stanford University.}}
\author[2]{Ekin Dogus Cubuk}
\author[1]{Surya Ganguli}
\author[2]{Samuel S. Schoenholz}
\affil[1]{Stanford University, Stanford, CA, USA}
\affil[2]{Google Research, Mountain View, CA, USA}

\begin{document}

	\maketitle
	
	\begin{abstract}
		\noindent
		Deep neural network classifiers partition input space into high confidence regions for each class. The geometry of these class manifolds (CMs) is widely studied and intimately related to model performance; for example, the margin depends on CM boundaries. We exploit the notions of Gaussian width and Gordon's escape theorem to tractably estimate the effective dimension of CMs and their boundaries through tomographic intersections with random affine subspaces of varying dimension. We show several connections between the dimension of CMs, generalization, and robustness. In particular we investigate how CM dimension depends on 1) the dataset, 2) architecture (including ResNet, WideResNet \& Vision Transformer), 3) initialization, 4) stage of training, 5) class, 6) network width, 7) ensemble size, 8) label randomization, 9) training set size, and 10) robustness to data corruption. Together a picture emerges that higher performing and more robust models have higher dimensional CMs. Moreover, we offer a new perspective on ensembling via intersections of CMs. Our code is on \href{https://github.com/stanislavfort/slice-dice-optimize/blob/main/cutting_planes_in_JAX.ipynb}{Github}.
	\end{abstract}

	\section{Introduction}
	Training neural networks to classify data is a ubiquitous and classic problem in deep learning. In $K$-way classification, trained networks naturally partition the space of inputs into $K$ types of regions, $S_k\subset R^D$, containing points that the network confidently predicts have class $k$. We call these regions \emph{class manifolds} (CMs) of the neural network. In this paper, we analyze the high-dimensional geometry of these CMs, focusing primarily on their \textit{effective dimension} that we define using the Gordon's escape through a mesh theorem \citep{gordon1988milman} and the concept of Gaussian width from high-dimensional geometry \citep{vershynin_2018}. 
	
	To estimate the dimension of these class manifolds, we perform constrained optimization on random $d$-dimensional sections (affine subspaces, which are $d$-dimensional generalizations of lines, planes etc) of input space to actively seek out regions that the neural network assigns to a target class with high confidence. Using optimization in this way allows us to beat the curse of dimensionality \citep{Bellman:DynamicProgramming} and find points in the input space that are unlikely to be discovered with other diagnostic techniques such as random sampling. Through a theoretical analysis of high-dimensional geometry, we link the success of such constrained optimization to the effective dimension of the target class manifold using the Gordon's escape through a mesh theorem \citep{gordon1988milman} and the concept of Gaussian width of a set \citep{vershynin_2018}. Using extensive experiments, we leverage this method to show deep connections between the geometry of CMs, generalization, and robustness. In particular we investigate how CM dimension depends on the dataset, architecture (including ResNet \cite{he2015deep}, WideResNet \citep{zagoruyko2017wide}, and the Vision Transformer \citep{dosovitskiy2020image}), random initialization, stage of training, class, network width, ensemble size, label randomization, training set size, and model robustness to data corruption. Together a picture emerges that well-performing, robust, models have class manifolds that have higher dimension than inferior models. As a corollary, we offer a unique geometric perspective on ensembling via intersections of CMs.
	\begin{figure}[!ht]
		\centering
		\includegraphics[width=0.45\linewidth]{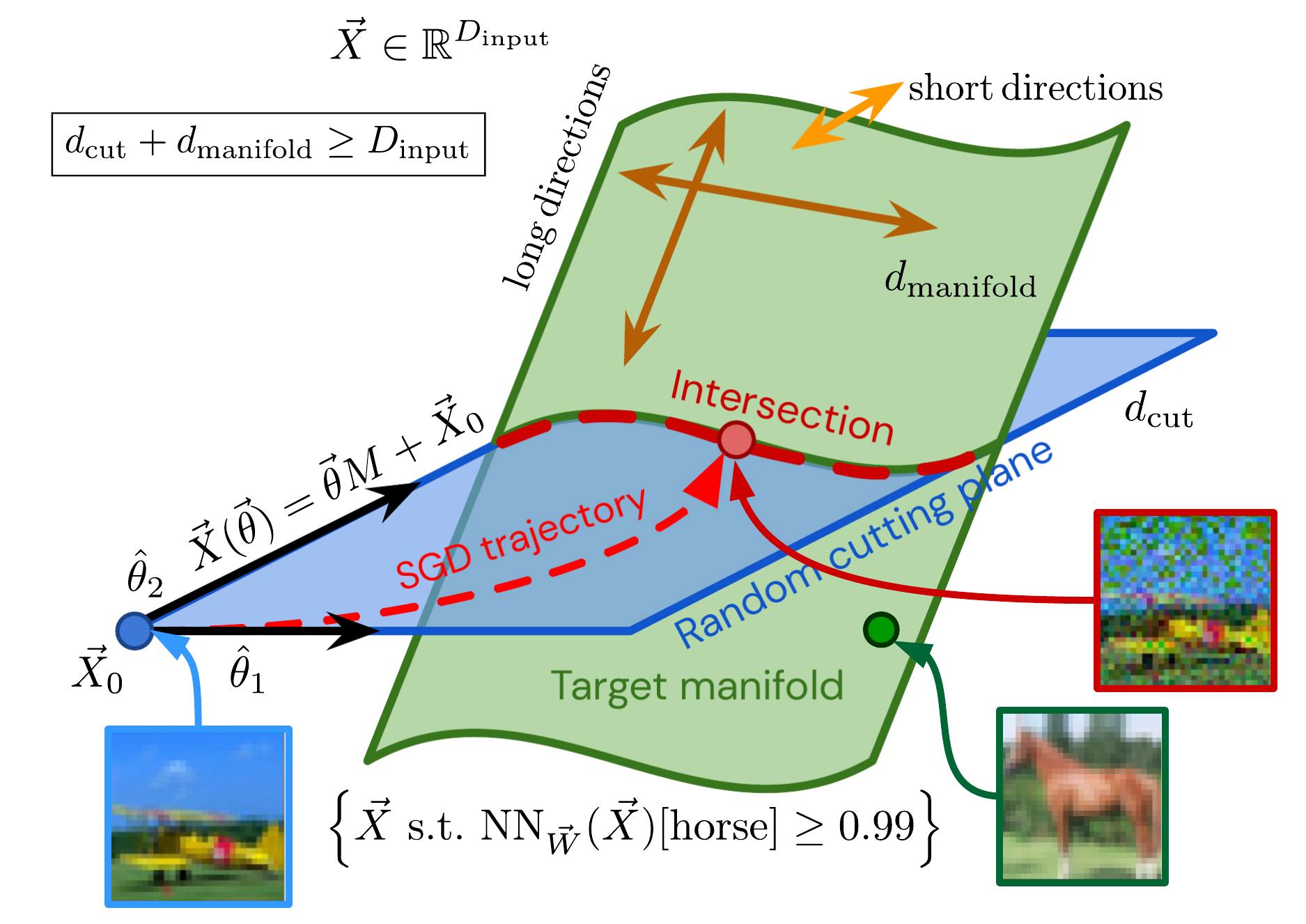}
		\raisebox{0mm}{\includegraphics[width=0.3\linewidth]{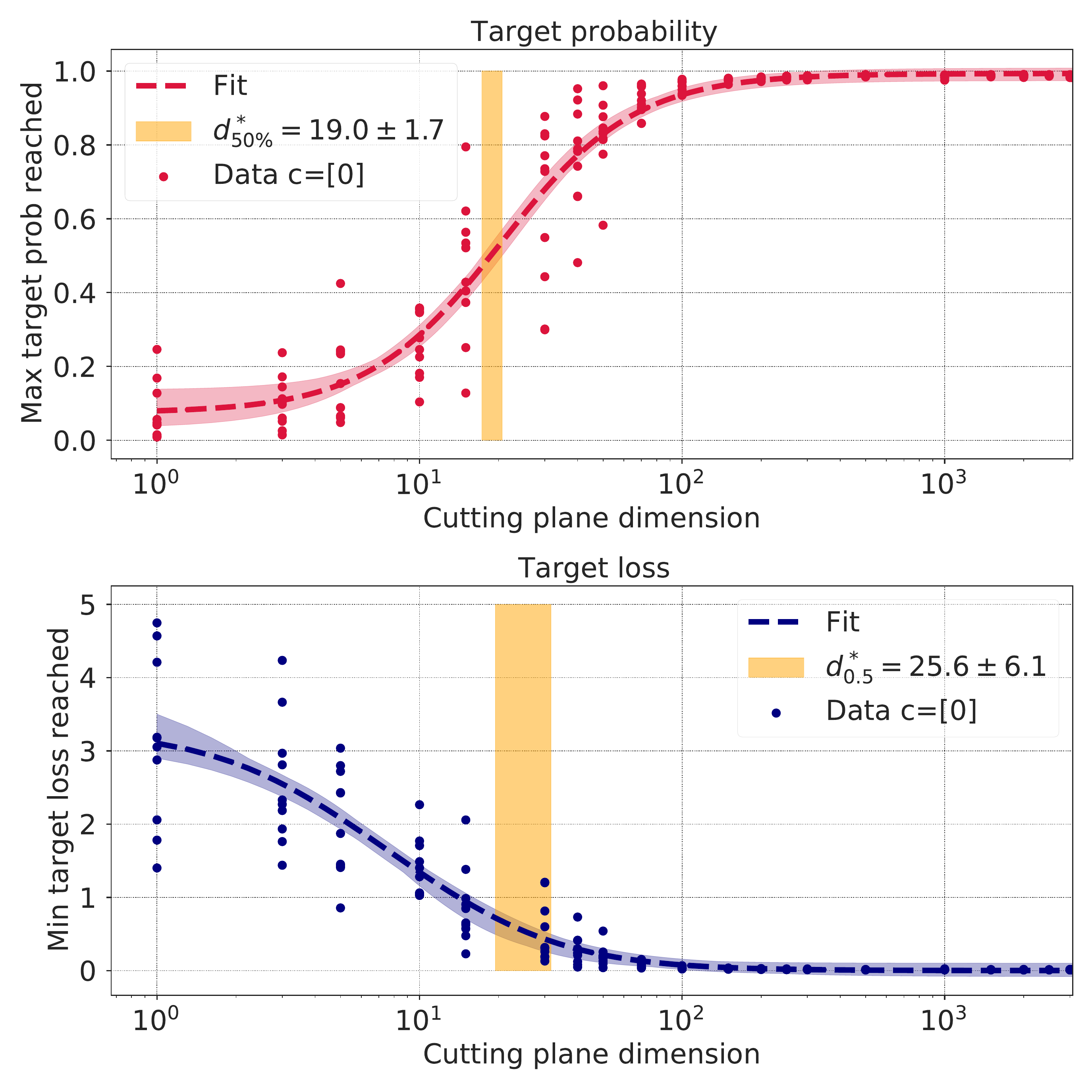}}
		\caption{An illustration of finding a point in the intersection between a random cutting plane of dimension $d_{\mathrm{cut}}$ (affine subspace) and a high-confidence class manifold (CM) of effective dimension $d_{\mathrm{manifold}}$. If the $d_{\mathrm{cut}} \gtrapprox D_{\mathrm{input}} - d_{\mathrm{manifold}}$, there likely exists an intersection between the two. We use optimization from a random point (image) $\vec{X}_0$ on the $d_{\mathrm{cut}}$ affine subspace to find a point in the intersection using gradient descent. The panels below show an example of the dependence of the maximum class probability and minimum loss reached vs. cut dimension $d_{\mathrm{cut}}$. The higher dimensional the cut, the less constrained the available images $\vec{X}$ are, and the more likely we are to find one of high class confidence.}
		\label{fig:how_to_cut}
	\end{figure}
	\begin{figure*}[ht]
		\centering
		\includegraphics[width=0.8\linewidth]{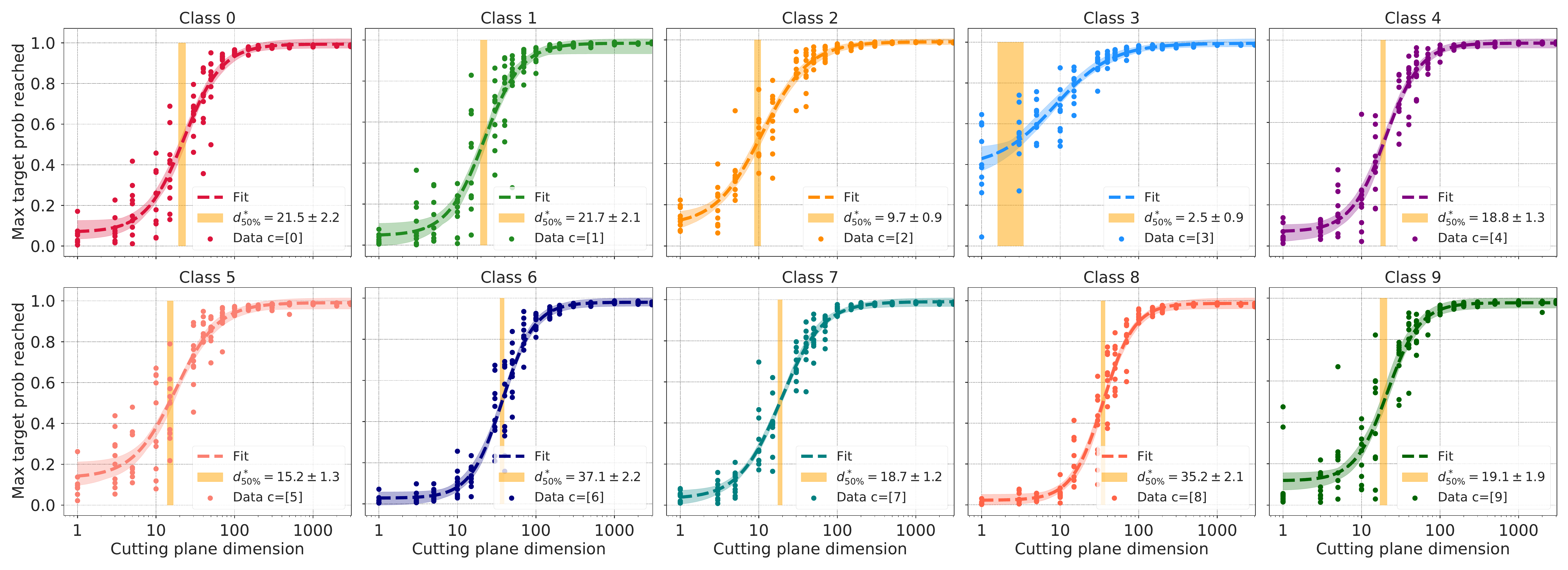}
		\caption{Maximum probability of single classes of CIFAR-10 reached on random cutting planes of dimension $d$ through optimization. The figure shows dependence of the probability of a single class of CIFAR-10 (y-axes) reached via optimization on random cutting affine subspaces of different dimensions (x-axes). The results shown are for a well-trained ($>90\%$ test accuracy) ResNet20v1 on CIFAR-10. The experiment for each dimension $d$ is repeated $10\times$ with random spans and offsets chosen from the training set. $d^*_{50\%}$ is extracted using a fit. The $d^*_{50\%} \ll 3072$, which implies that the class manifolds are surprisingly high dimensional ($3072 - d^*_{50\%}$) (\textit{all in excess} of $3000$).}
		\label{fig:single_class_prob_curves_cifar10}
	\end{figure*}
	
	\textbf{Related work.} There has been significant research into understanding linear regions of neural networks, both trained and untrained. \citet{montfar2014number} studied the number of linear regions in deep neural networks, \citet{raghu2016expressive} looked at their expressive power with depth, while \citet{serra2017bounding, novak2018sensitivity} tried to bound and count them. \citet{hanin2019deep} showed that deep $\mathrm{ReLU}$ networks have surprisingly few activation patterns, and \citet{hanin2019complexity} did the same for the linear regions in the input space. The spectral properties of neural nets were studied in \citet{rahaman2018spectral}, and the stiffness of the functional approximations defined through gradient alignment was coined in \citet{fort2019stiffness}. \citet{balestriero2018mad} and \citet{balestriero2019geometry} use splines to understand class bounderies. While revealing interesting aspects of neural network input space and activations space behavior, the methods used so far have not been able beat the curse of dimensionality -- they have stayed local, and analyzed either one- or two-dimensional sections of input space.  While our method makes a global estimate of the dimension, local methods based on Maximum Likelihood Estimate pioneered in \citet{levina2005maximum} sparked a fruitful research direction, for example continued by \citet{ma2018characterizing} and their application to adversarial examples.
	
	The exploration of constrained optimization on random, $d$-dimensional planes in the weight space was employed successfully in \citet{li2018measuring} to estimate the intrinsic dimension of loss landscapes. \citet{Fort_2019} extended this analysis geometrically, and \citet{fort2019large} used this and other observations to build a geometric model of the low-loss basins weight-space basins.
	
	Another closely related area concerns adversarial examples and robustness. \citet{szegedy2013intriguing} first noted that there exist points in input space very close to test examples that are mispredicted by neural networks, suggesting CMs of different classes can come very close to each other. \citet{gilmer2018adversarial} showed that the existence of adversarial examples is related to the dimensionality of input space and the accuracy of the classifier. \citet{ford2019adversarial} further link this interplay between dimension, generalization, and adversarial robustness to more general corruption robustness. In a similar spirit \citet{hadi_robustness2019} produce more robust models by convolving neural networks with Gaussian noise in input space. \citet{ovadia2019trust} explore model uncertainty in general. Whereas these studies are local, the techniques discussed in this paper are primarily concerned with global properties of CMs.
	
	\section{Methods}
	
	We seek to determine the effective dimension of class manifolds (CMs). To that end, consider a neural network whose last layer is a $\mathrm{softmax}$ yielding normalized probabilities for a given input, $\vec p(\vec X)$. We define a class manifold for class $k$ as the pre-image,
	$S_k =  \vec p^{-1}(\{p_k > p_{\mathrm{threshold}}\})$ for some confidence threshold $p_{\mathrm{threshold}}$. 
	We seek to identify the effective dimension of $S_k$ by introducing \emph{the Subspace Tomography Method} (see also Fig.~\ref{fig:how_to_cut}):
\vspace{0.1cm}
\fbox{
			\parbox{0.96\linewidth}{
			\textbf{The Subspace Tomography Method:} For a neural network ($\mathrm{NN}$) mapping input $\vec{X} \in \mathbb{R}^D$ into probabilities $\vec{p}(\vec X) \in \mathbb{R}^C$, take a random $d$-dimensional affine subspace defined by orthonormal basis vectors given by rows of $M \in \mathbb{R}^{d \times D}$ and a point $\vec{X}_0 \in \mathbb{R}^D$ (from the training set in our case). Inputs in this subspace are parameterized by $\vec\theta\in\mathbb R^d$ as $\vec{X}(\vec{\theta}) = \vec{\theta} M + \vec{X}_0$. Given a target probability vector $\vec{p}_{\mathrm{target}}$, we seek to optimize the cross entropy loss, $\mathcal{L} (\vec p(\vec X(\vec{\theta})), \vec p_{\mathrm{target}})$ with respect to $\vec{\theta}$. This will identify points constrained to the affine subspace $(M,\vec{X}_0)$ that have probabilities as close as possible to $p_{\mathrm{target}}$. We study the dependence of $\mathcal{L}_{\mathrm{min}}$ and $\vec{p}_{\mathrm{max}}$ (the loss and probability after optimization) over many repetitions of the procedure on the cut dimension $d$. We show this analysis estimates the effective codimension of the pre-image of $p_{\mathrm{target}}$: $\{ \vec{X} \,\mathrm{s.t.}\,\vec p(\vec{X}) = \vec{p}_{\mathrm{target}} \}$ by observing for which $d_{\mathrm{cut}}$ the expected $p_{\mathrm{max}}$ reaches a threshold.
		}
		\label{box:subspace_tomography_method}
	}

	As discussed in the summary box, we use the cross entropy loss $\mathcal L(p(\vec X), \hat p) = -\hat p\cdot\log[\vec p(\vec X)]$ between the target probability vector $\vec{p}_{\mathrm{target}}$ and the output of the network to reach the intersection. We use \verb|Adam| \citep{kingma2017adam} to minimize $\mathcal{L}$ with respect to $\vec{\theta}$, starting from $\vec{\theta}_0 = \vec{0}$, which corresponds to an initial {\it random} input $\vec{X}(\vec{\theta}_0) = \vec{X}_0$. We choose this $\vec{X}_0$ such that it is not of any of the target classes whose dimension we are trying to measure, as discussed in Section \ref{sec:method_classes}. We found no effect of choosing $\vec{X}_0$ from different distributions, and decided to use the training set.
	Through optimization, we take $\vec{\theta}_0 \to \vec{\theta}_{\mathrm{min}}$.  The $\vec{\theta}_{\mathrm{min}}$ defines an optimized input $\vec{X}_{\mathrm{min}} = \vec{\theta}_{\mathrm{min}} M + \vec{X}_0$ and corresponding output $\vec{p}_{\mathrm{max}} = \vec{p}(\vec{X}_{\mathrm{min}})$ that is as close as possible to $\vec{p}_{\mathrm{target}}$ while confining $\vec{X}$ to the random affine subspace (cut) defined by $(M,\vec{X}_0)$. As a technical detail, we discuss the weak effect of sparsity of $M$ in Fig. \ref{fig:plane_alignemnt_effect}. 
	
	The optimization thus starts with a tuple $(\mathrm{NN},d,M,\vec{X}_0)$ and maps it to the final probability vector $\vec{p}_{\mathrm{max}}$ and the associated $\mathcal{L}_{\mathrm{min}}$. By analyzing the dependence of $\vec{p}_{\mathrm{max}}$ and $\mathcal{L}_{\mathrm{min}}$ on the dimension $d$ of the cut we can estimate the effective dimension of the pre-image in input space of a region around $\vec{p}_{\mathrm{target}}$ in output space (Fig. \ref{fig:how_to_cut}).
	
	\citet{lotterysubspaces} use the Subspace Tomography Method to explore the manifold of solutions in the weight space by looking for low-loss parameter configurations on affine subspaces of various types.
	
	\subsection{Class manifolds (CMs) and multi-way class boundary manifolds (CBMs)}
	\label{sec:method_classes}
	There are several interesting choices of $\vec{p}_{\mathrm{target}}$. Consider $\vec{p}_{\mathrm{target}} = (0,0,\dots,1,\dots,0)$, a 1-hot vector on a single class $k$. The pre-image of $\vec{p}_{\mathrm{target}}$ is the CM $S_k$, the set of points in the input space that map to high-confidence class $k$ predictions. The cutting plane method allows us to estimate the effective co-dimension of $S_k$ by computing the dimension $d^*$ at which we reliably obtain a $\vec{p}_{\mathrm{max}}$ whose $k$'th component is close to $1$ within some tolerance (Fig. \ref{fig:single_class_prob_curves_cifar10}). More precisely, by choosing a threshold $p^*$, we are detecting the super-level set of inputs $\{ \vec{X} \in \mathbb{R}^D \,\,\mathrm{s.t. }\,\,\ \vec{p}(\vec{X})[k] \ge p^* \}$ (see e.g. Fig.~\ref{fig:single_class_prob_curves_cifar10}).
	
	The cross entropy loss formulation allows us to also study regions that lie in between classes. For example, by setting $\vec{p}_{\mathrm{target}} = (\frac{1}{2},\frac{1}{2},0,\dots,0)$, our optimization finds regions of input space that lie on a class boundary manifold (CBM) between classes 0 and 1. We can even find multi-way CBMs. For example, a three-way CBM between classes 0,1, and 2 corresponds to $\vec{p}_{\mathrm{target}} = (\frac{1}{3},\frac{1}{3},\frac{1}{3},0,\dots,0)$. At the extreme, we can study the region where all classes have equal probability by setting  $\vec{p}_{\mathrm{target}} = (\frac{1}{C},\frac{1}{C},\dots,\frac{1}{C})$, where $C$ is the number of classes. The subspace tomography method, therefore, allows us to study the intertwined geometry of multiple CMs and their boundaries. The nature of this geometry is deeply linked with generalization, via the margin, and adversarial examples. See Fig.~\ref{fig:multi_class_prob_and_loss_curves_cifar10} for results on multi-way CBMs.
	
	\subsection{Extracting the critical cutting plane dimension and CM co-dimension $d^*_{\mathrm{50\%}}$}
	Given a particular class target vector $\vec{p}_{\mathrm{target}}$ (e.g. $\vec{p}_{\mathrm{target}} = (1,0,\dots,0)$ corresponding to the CM $S_k$ with $k=1$), we perform the subspace tomography experiment multiple times for random $\vec{X_0}$ and $M$ (both randomly chosen for every experiment) for a sweep of different values of $d$. For each random draw of $M$ and $\vec{X_0}$, we obtain a final probability vectors $\vec{p}_{\mathrm{max}}$ as a function of cutting plane dimension $d$, as shown e.g. in Fig.~\ref{fig:how_to_cut} and \ref{fig:single_class_prob_curves_cifar10}. When targeting a single class manifold $S_k$ we plot the $k$'th component $\vec{p}_{\mathrm{max}}[k]$. For small values of $d$, the affine cutting plane will not intersect the target manifold, $S_k$, and $\vec{p}_{\mathrm{max}}$ will be far from $\vec p_{\mathrm{target}}$. For large dimensions, e.g. $d=D$, the subspace is now the full space of inputs, and we can always find a point on the plane such that $\vec{p}_{\mathrm{max}} \approx \vec p_{\mathrm{target}}$. For intermediate values of $d$, the $k$'th component of $\vec{p}_{\mathrm{max}}$ will gradually increase with $d$ in expectation. To extract a single cutting plane dimension from this data we 1) fit an empirical curve to the data (Equation \ref{eq:fit}; typically a good fit), 2) use the mean and covariance of the fitting parameters to obtain a distribution of valid fitting functions, and 3) extract the range of values of $d^*$ where these functions cross a threshold probability, often $p = 50\%$. We call this value $d^*_{\mathrm{50\%}}$; in some cases we use thresholds of $25\%$ and $75\%$, and in principle we can choose whichever we like, understanding that it measures the appropriate superlevel set and we note that in the figures. This cutting plane dimension $d^*$ is the \textit{effective co-dimension} of the CM $S_k$. Thus the \textit{effective dimension} of the CM $S_k$ is $D-d^*$ (as derived in Section \ref{sec:theory}).

	\section{A theory for estimating class manifold dimension through the Tomographic Subspace Method (TSM)}
	\label{sec:theory}
	
	We begin with a simple theoretical description of our method for the case of affine subspaces, after which we will consider the case of realistic CMs. 
	
	\textbf{High-level description.} Two affine subspaces of dimensions $d_A$ and $d_B$ generically intersect provided that their dimensions add to at least the dimension of the ambient space $D$ they are embedded in, $d_A + d_B \ge D$.
	
\begin{figure}[ht]
		\centering
		\includegraphics[width=0.85\linewidth]{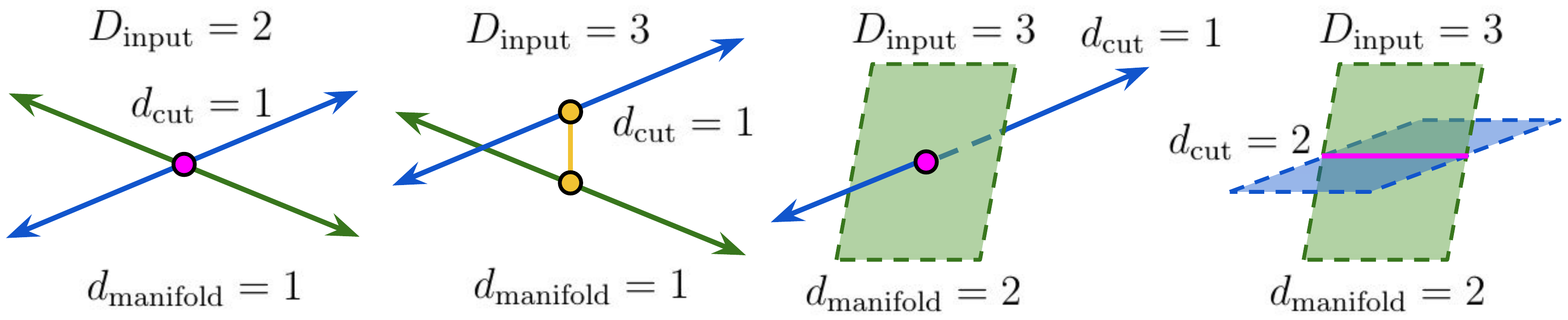}
		\caption{An illustration of the way two affine subspaces of dimensions $d_A$ and $d_B$ can intersect. If the dimensions add up to at least the ambient dimension $D$, $d_A + d_B \ge D$, the subspaces likely intersect, otherwise they typically do not. If we know $D$ and $d_A$, we can use the reliable existence of an intersection to bound $d_B \ge D - d_A$. An equivalent result holds beyond affine subspace.}
		\label{fig:cut_options}
	\end{figure}

	In algebraic geometry, this statement is known as \textit{dimension counting}, and is equivalent to the statement that the co-dimensions of subspaces are at most additive under intersection \citep{bourbaki1998algebra} (recall that the co-dimension of a subspace of dimension $d$ in a space of ambient dimension $D$ is $D-d$). An illustration of what such intersections can look like for $D=2$ and $D=3$ are shown in Fig.~\ref{fig:cut_options}. 
	
	If we know that there reliably exists an intersection, we can use this fact to bound $d_B \ge D - d_A$. The same intuition carries over to a situation where an affine subspace $A$ of dimension $d_A$ intersects a generic manifold $B$ of effective dimension $d_B$. Our Tomographic Subspace Method uses constrained optimization on randomly chosen affine subspaces, $A$, to measure the lowest dimension, $d^*$, at which $A$ reliably intersects $B$, a class manifold in the input space. This may therefore be used to bound the dimension of $B$ as $d_B \ge D - d^*$. By replacing the linear algebra dimension of the subspace $B$ with the effective dimension of the class manifold, the condition for intersection remains unchanged.

	%
	
\subsection{Gaussian width and the diameter of a set}
Our goal is to study the class manifolds (CMs) and class boundary manifolds (CBMs) in the space of inputs of deep neural networks. For a $\mathrm{NN}: \vec{X} \in \mathbb{R}^D \to \vec{p} \in \mathbb{R}^C$ mapping inputs of dimension $D$ to class probabilities of dimension $C$, the manifolds in question are the pre-images of a particular target output $p_{\mathrm{target}}$: $\{ \vec{X} \,\mathrm{s.t.}\,\vec p(\vec{X}) = \vec{p}_{\mathrm{target}} \}$.
	\begin{wrapfigure}{r}{0.5\textwidth}
		\centering
		\includegraphics[width=0.8\linewidth]{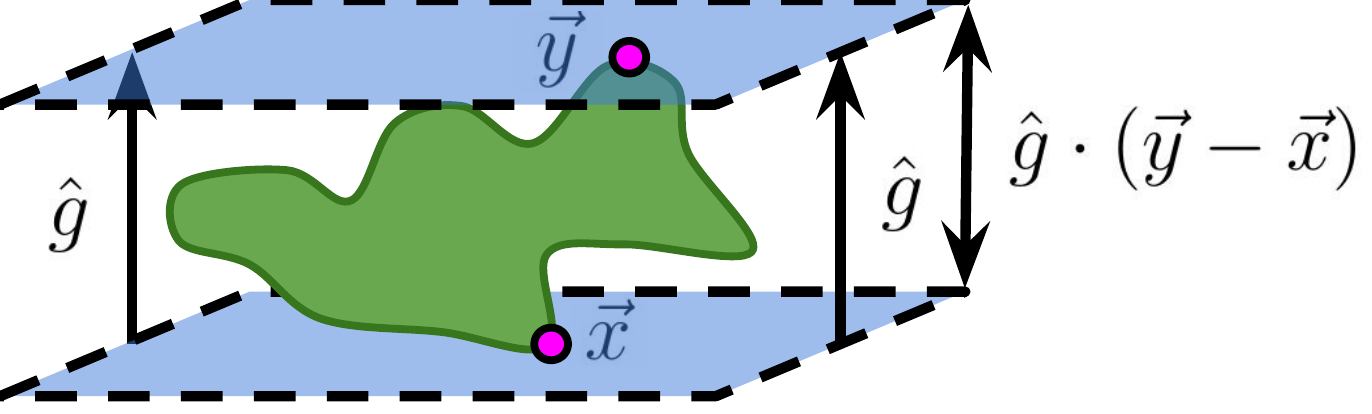}
		\caption{An illustration of measuring the Gaussian width of a set $T$ (in green) in a direction $\hat{\vec{g}}$ by identifying $\vec{x},\vec{y} \in T$ in $\max_{\vec{x},\vec{y} \in T} \hat{\vec{g}} \cdot (\vec{y} - \vec{x})$. The expectation width using random vectors $\vec{g} \sim \mathcal{N}(0,1)^D$ instead of $\hat{\vec{g}}$ is a half of the Gaussian width $w(T)$. Intuitively, it is the characteristic extent of the set $T$ over all directions rescaled by a factor between $D/\sqrt{D+1}$ and $\sqrt{D}$.}
		\label{fig:gaussian_width}
	\end{wrapfigure}
Our method uses the empirically estimated probability of an intersection of an affine subspace with a manifold to measure the dimension of the manifold. The probability of an intersection therefore depends on the extent of the manifold in different directions. For an affine subspace of dimension $n$, we have $n$ dimensions of length $\infty$, and $D-n$ dimensions of length $0$. For a generic set, however, the situation is more complicated.

To estimate the effective dimension (which is equal to the statistical dimension \citep{Amelunxen_2014}) of a subset $T \subseteq \mathbb{R}^D$, we need to measure its Gaussian width as defined in \citet{vershynin_2018}. 
We denote the Gaussian width of the set $T$ by $w(T)$. In words, $w(T)$ is defined to be half of the expected diameter of $T$ as measured over all directions and rescaled by the length of a random vector $\vec{g} \in \mathcal{N}(0,1)^D$. The expected length of this vector is bounded by $D / \sqrt{D+1} < \mathbb{E}(|\vec{g}|_2) < \sqrt{D}$ \citep{mixon_2014}, and as $D \to \infty$, $|\vec{g}| \to \sqrt{D}$. Along a direction $\hat{g}$, the width of the set $T$ is $\max_{\vec{x},\vec{y} \in T} \left ( \hat{g} \cdot (\vec{x} - \vec{y}) \right )$. Therefore, mathematically, the Gaussian width is defined as
	\begin{equation}
	\begin{split}
	w(T) = & \frac{1}{2} \mathbb{E}_{\vec{g} \sim \mathcal{N}(0,1)^D}{ \max_{\vec{x},\vec{y} \in T} \left ( \vec{g} \cdot (\vec{x} - \vec{y}) \right )} \\ = & \mathbb{E}_{\vec{g} \sim \mathcal{N}(0,1)^D}{ \max_{\vec{x} \in T} \left ( \vec{g} \cdot \vec{x} \right )} \, .
	\end{split}
	\label{eq:gaussian_width}
	\end{equation}
By contrast, the diameter of the set is its maximum extent over all directions
\begin{equation}
\mathrm{diam}(T) = \max_{\vec{g} \sim \mathcal{N}(0,1)^D} \max_{\vec{x},\vec{y} \in T} \left ( \vec{g} \cdot (\vec{x} - \vec{y}) / |\vec{g}|_2\right ) \, .
\end{equation}

	\subsection{Effective and statistical dimension}
    The linear algebra concept of a dimension of a subset $T$ is the smallest dimension of an affine subspace that contains $T$. This definition is very brittle -- an infinitesimal perturbation to a single point in $T$ can change the resulting dimension \citep{vershynin_2018}. In high-dimensional geometry, \textit{effective dimension} \citet{vershynin_2018} and \textit{statistical dimension} \citet{Amelunxen_2014} are both robust and can be estimated using our Subspace Tomography Method.

\textbf{Gordon's escape through a mesh theorem.}
As described above, when the target manifold is affine, we know the exact condition for there to exist an intersection with a random affine subspace: their dimensions must add up to at least the dimension of the ambient space, $d_{\mathrm{cut}} + d_{\mathrm{target}} \ge D$. For generic target subsets, $T$, the condition turns out to be very similar. To show that, we will use the Gordon's escape through a mesh theorem \citep{gordon1988milman, mixon_2014, Amelunxen_2014}.

A complication, however, is that the theorem is defined for subsets of the unit sphere $S \subseteq \mathbb{S}^{D-1}$ centered on the point $\vec{X}_0$ contained in the cutting plane, rather than a generic subset $T \subseteq \mathbb{R}^D$. We resolve this by noticing that were we to project $T$ to the surface of the unit sphere as $\mathrm{proj}_{\vec{X}_0}(T) = \{\vec{X}_0 + (\vec{x}-\vec{X}_0) / ||(\vec{x}-\vec{X}_0)||_2 \,\,\mathrm{for}\,\,\forall \vec{x} \in T\}$, for any cutting plane $A$ passing through $\vec{X}_0$ the probabilities of intersection are exactly the same, 
\begin{equation}
    \mathrm{Pr} \left ( A_{\vec{X}_0} \cap T \neq \emptyset \right ) = \mathrm{Pr} \left (A_{\vec{X}_0} \cap \mathrm{proj}_{\vec{X}_0}(T) \neq \emptyset \right ) \, .
\end{equation} Since $\mathrm{proj}_{\vec{X}_0}(T) \subseteq \mathbb{S}_{\vec{X}_0}^{D-1}$, we will refer to $S = \mathrm{proj}_{\vec{X}_0}(T)$ and derive the result below, noting that the same holds for $T$. The effective dimension measured in this way will therefore be $\vec{X}_0$ dependent. In practice, we marginalize over different values of $\vec{X}_0$ to produce a consistent estimate of effective dimension.
	
The Gordon's escape through mesh theorem allows us to bound the probability that a linear subspace $A$ of dimension $d$, and co-dimension $k = D-d$, will not intersect the subset $S$ in terms of its Gaussian width $w(S)$.
\begin{equation}
	\mathrm{Pr}\left ( A \cap S = \emptyset \right ) \ge 1 - \frac{7}{2} e^ { - \frac{1}{18}\left ( a_k - w(S) \right )^2 } \, ,
	\label{eq:gordon}
\end{equation} 
where $k / \sqrt{k+1} < a_k < \sqrt{k}$ and the bound is valid only for $w(S) < a_k$. Since we typically have $k \gg 1$ we can assume $a_k = \sqrt{k}$. 

The probability of a miss goes down up to the point where $w(S) = a_k$, which we will use to define the effective dimension. Since $a_k \approx \sqrt{D-d}$, this corresponds to $w^2(S) = D-d$. Comparing this to the affine subspace case, we see that $w^2(S)$ now acts as the effective dimension of the target set $T$ whose projection $S = \mathrm{proj}(T)$ we're studying. 
\begin{equation}
\boxed{
    d_{\mathrm{effective}}(T \in \mathbb{R}^D, \vec{X}_0) = w^2 \left ( \mathrm{proj}_{\vec{X}_0}(T) \in \mathbb{S}_{\vec{X}_0}^{D-1} \right )
}
\label{eq:effective_dimension}
\end{equation}

\subsection{Affine subspaces as a corollary and numerical experiments}
In a way, the Gaussian width allows us to count the number of \textit{long} directions of the set $T$ as compared to the distance of $T$ from the origin of the cutting plane. To help build some intuition, we will now apply Gordon's escape through the mesh theorem to the case of an affine target space considered above. Imagine an $n$-dimension affine subspace $T$; as described above, such a space is characterized by $n$ dimensions that are infinite in extent and $D - n$ dimensions that have no extent at all. The projection $\mathrm{proj}(T)$ will wrap around an angle $\pi$ of the unit sphere along the $n$ axes of infinite extent, and will have 0 extent along the others. Therefore $w(S) = \sqrt{n}$ (assuming $D, n \gg 1$). Using Eq.~\ref{eq:effective_dimension}, $d_{\mathrm{effective}} = n$, recovering the dimension of the affine subspace we chose to use.

	\begin{figure}[ht]
		\centering
		\includegraphics[width=0.3\linewidth]{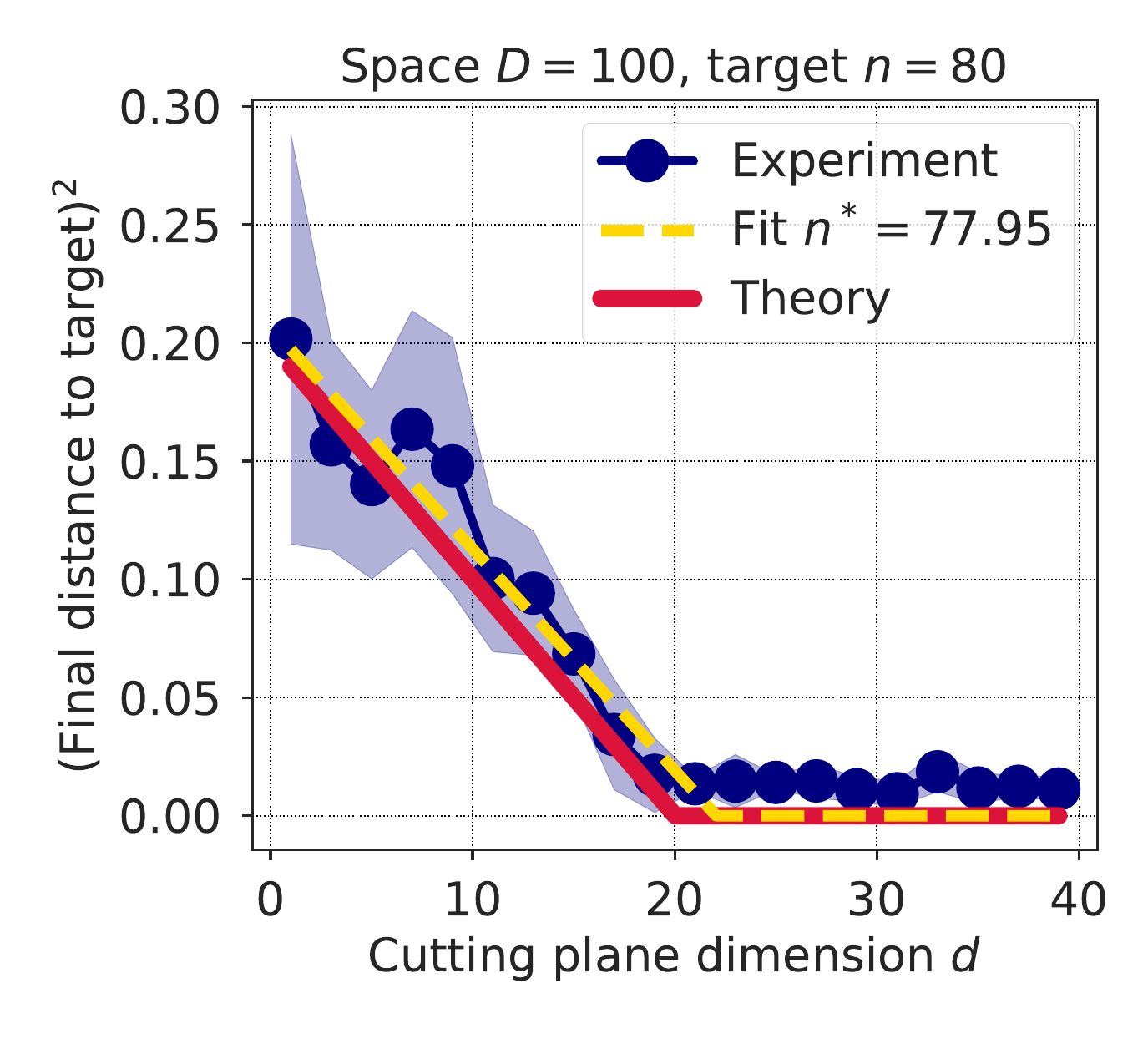}
		\includegraphics[width=0.27\linewidth]{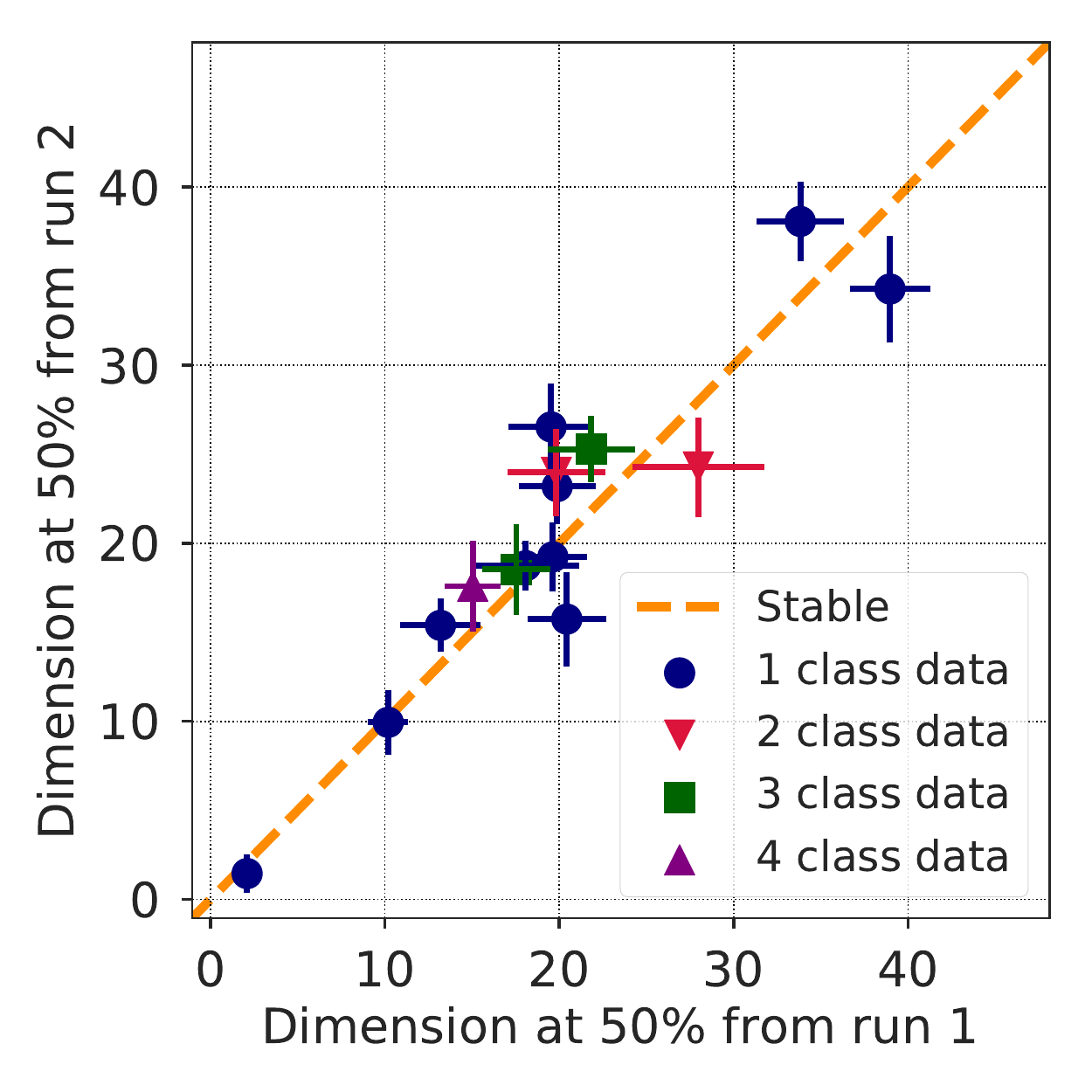}
		\caption{\textbf{Left panel:} Distance between random affine subspaces -- numerical experiments vs theory from Eq. \ref{eqn:distance_analytic_scaling}. The figure compares the distance between two subspaces of dimensions $d$ \& $n$ in a $D$-dimensional space. The numerical experiment in JAX is in blue, theory in red, and numerical fit in yellow. \textbf{Right panel:} Comparison of the cutting plane dimension needed to get $50\%$ of the target class ($d^*_{50\%}$) for two independently initialized and trained ResNets on CIFAR-10, showing the stability of our method to reinitialization and retraining. 
		}
		\label{fig:toy_model_comparison}
		\label{fig:two_inits_comparison}
	\end{figure}
	
	In our subspace tomography method, we control the dimension $d_A$ of a randomly chosen cutting affine subspace and use optimization constrained to it to find an intersection with a class manifold in order to estimate its effective dimension $d_B$. The dimension $d_A = d^*_{50\%}$ where we can first reliably find a $50\%$ probability image of the target class will be an estimate of the \textit{codimension} of such a CM. An estimate of the dimension of the CM will therefore be $d_B = D - d^*_{50\%}$.

		In Sec.~\ref{sec:analytics} we analytically derive the expected closest distance between two such affine subspaces. The result for $d_A + d_B \ge D$ is exactly 0 (they itersect), while for $d_A + d_B < D$ the $\mathbb{E}\left [ l(A,B) \right ] \propto (\sqrt{D - d_A - d_B}) / \sqrt{D}$. To compare this analytic result to reality, we ran a numerical experiment using automatic differentiation in JAX \citep{jax2018github} where we generated random affine subspaces of different dimensions and measured their closest approach using optimization to locate the point of closest approach (or intersection). An example is shown in Fig.~\ref{fig:toy_model_comparison}.

	\section{Experiments}
	\begin{figure}[ht]
			\centering
			\includegraphics[width=0.6\linewidth]{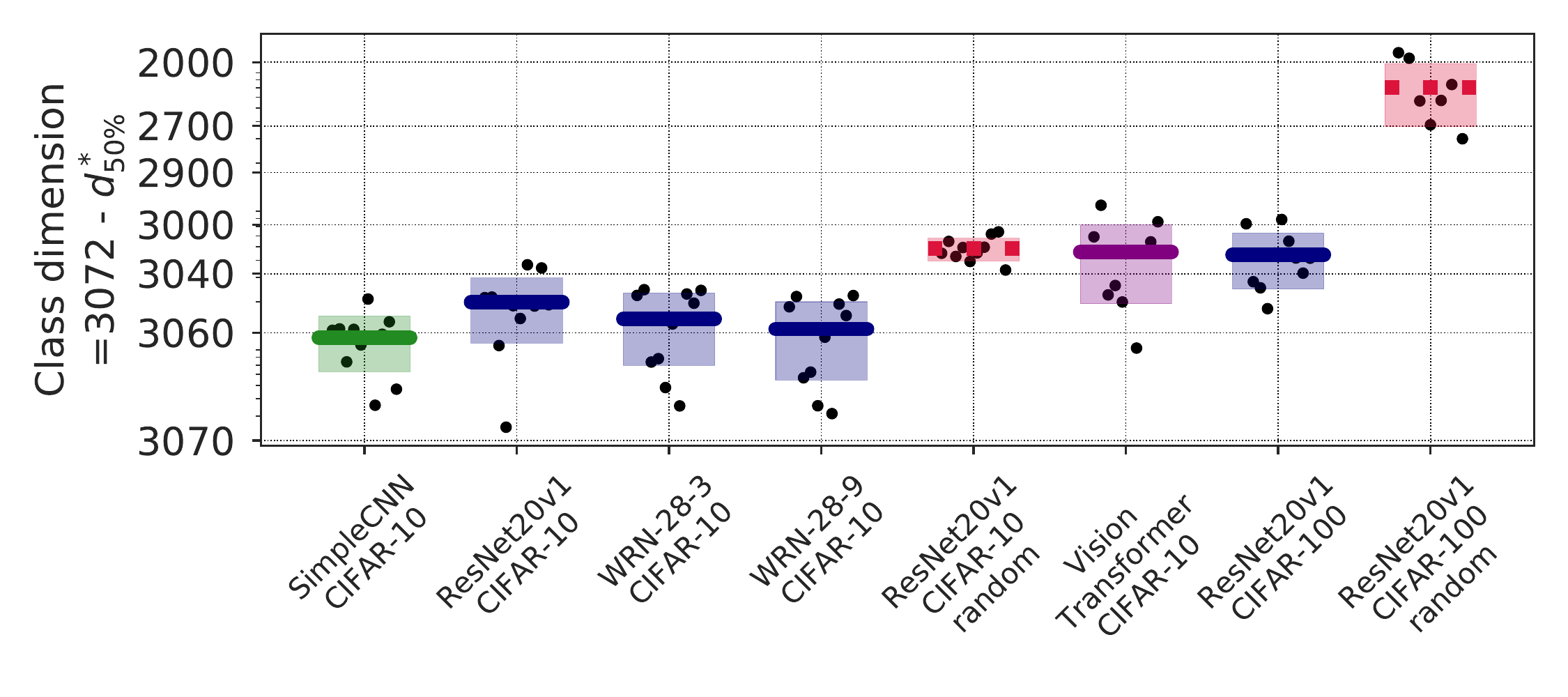}
			\caption{Comparison of the class manifold dimensions ($3072 - d^*_{50\%}$) for SimpleCNN, ResNet, WideResNet and Vision Transformer on CIFAR-10/100 with real and \textit{randomized} labels. Label randomization decreases the class manifold dimension. The dimensions of the learned class manifolds are very high as compared to dimensions of the data alone (Sec.~\ref{sec:dataset_dimension}).}
			\label{fig:models_plot}
		\end{figure}
	We now present our experiments using the tomographic subspace method to measure the dimension of CMs and CBMs, and to make connections to generalization and robustness. The details of the architectures, datasets and precise training procedures are in the Appendix Sec.~\ref{sec:training_details}. The majority of our experiments are done with a standard ResNet20v1 \citet{He_2016} and WideResNet on CIFAR-10 and CIFAR-100. To see how architecture-dependent our conclusions were, we also include results from the Vision Transformer model \citet{dosovitskiy2020image}, pretrained on ImageNet \citep{imagenet_cvpr09}, of a radically different design. For the cuts, we choose the random starting point $\vec{X}_0$ from the train set, making sure it is of a different class than contained in the target vector $\vec{p}_{\mathrm{target}}$. 

	\subsection{Re-initialization and re-training stability.}
	If the class manifold dimension is to be seen as a robust property, the results should be stable under reinitialization and retraining of a model. We verified that that is the case, as shown in Fig.~\ref{fig:two_inits_comparison}, comparing the $d^*_{50\%}$ dimensions extracted from single class regions of CIFAR-10, as well several regions between 2, 3 and 4 classes. The results are consistent between the 2 runs.

	\subsection{Single class manifolds.}
	The main object of interest for us are the high-confidence single class manifolds. To be precise, we study the supersets of class probability of a class $k$ above a threshold, most often $50\%$ to guarantee that $\mathrm{argmax}(\vec{p}) = k$. Given the continuity of the $\vec{p}(\vec{X})$, the supersets corresponding to higher thresholds will be subsets of the lower thresholds. We present our results for a well-trained ResNet20v1 on CIFAR-10 in Fig.~\ref{fig:single_class_prob_curves_cifar10}, and for CIFAR-100 in Fig.~\ref{fig:single_class_dim_vs_test_acc_resnet_cifar100}, for a SimpleCNN on CIFAR-10 in Fig.~\ref{fig:single_class_dim_vs_test_acc_simplecnn_cifar10}, and Vision Transformer in Fig.~\ref{fig:single_class_prob_curves_vision_transformer_cifar10}. The results show that the $d^*_{50\%}$ (dimension of the cutting plane) is $\ll$ the dimension of the input, therefore the class manifold dimension is very high (summary in Fig.~\ref{fig:models_plot}), close to the full 3072 dimensions for CIFAR (compared to small estimates of the dimension of the data itself, Sec.~\ref{sec:dataset_dimension}).

	\subsection{Class boundary manifolds between multiple classes.}
\begin{figure*}[ht]
		\centering
		\includegraphics[width=1.0\linewidth]{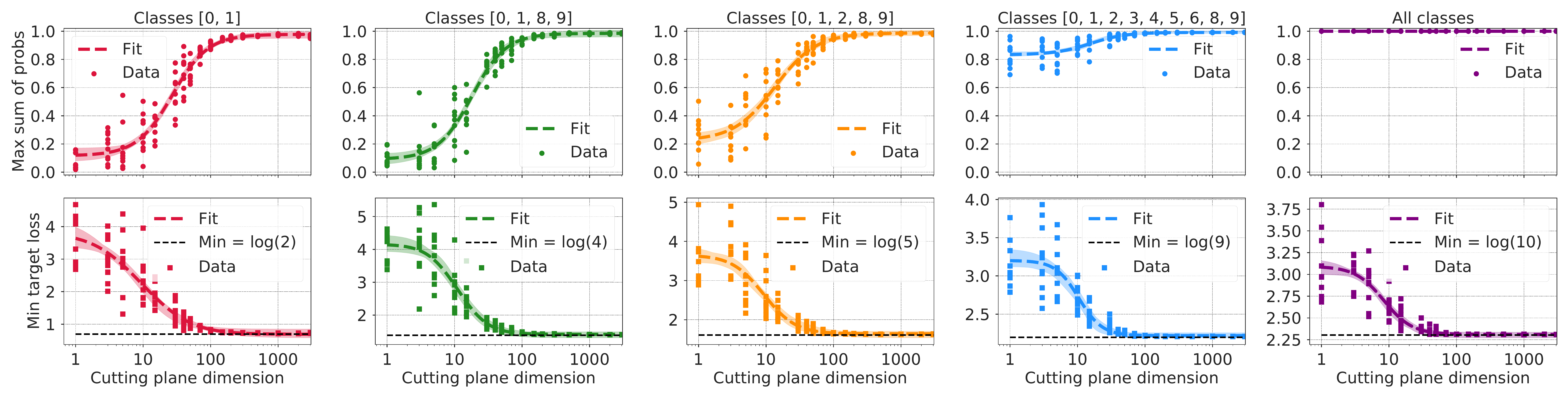}
		\caption{Maximum probability and minimum loss (y-axes) of in-between-classes regions of CIFAR-10 reached on cutting planes of dimension $d$ (x-axes).
			The results shown are for a well-trained ($>90\%$ test accuracy) ResNet20v1 on CIFAR-10. Each dimension $d$ is repeated $10\times$ with random planes and offsets (training examples of different than target classes). The last column shows the results for the $10$-class region where the network assigns equal probability to each class. The small spread of results for a given $d$ shows that the local difference in dimension are small and that we can estimate it well globally with our cutting plane method.}
		\label{fig:multi_class_prob_and_loss_curves_cifar10}
	\end{figure*}

	As described in Sec.~\ref{sec:method_classes}, our method allows us to study the dimension of boundary manifolds between multiple classes. We show results for a well trained ResNet20v1 on CIFAR-10 ($>91\%$ test accuracy) for several selected sets of classes in Fig.~\ref{fig:multi_class_prob_and_loss_curves_cifar10}. In particular, we look at the region in between all 10 classes, where the network is equally uncertain about all. There, we primarily focus on the loss (Sec.~\ref{sec:theory}) in the bottom row of Fig.~\ref{fig:multi_class_prob_and_loss_curves_cifar10}, since the probability always sums up to $1$.
	
	\subsection{Training on random labels.}
	Due to the structure of the training data and the neural network prior, we expect the learned class manifolds to inherit a lot of structure from both. To disentangle the role of the class label, we trained a ResNet20v1 on CIFAR-10 with randomly reshuffled labels. As shown in \citet{zhang2017understanding}, we can reach $100\%$ training accuracy on random labels with a network of high enough capacity. However, as shown in Fig.~\ref{fig:two_inits_comparison} and \ref{fig:permuted_single_class_resnet_cifar10} the class manifolds learned a significantly higher $d^*_{\mathrm{50\%}}$ and therefore smaller dimension than the ones corresponding to the semantically meaningful labels. Since these models completely fail to generalize, this result is consistent with the hypothesis that generalization and class manifold dimension are intimately related.
	
	\subsection{The effect of training set size.}
\begin{figure}[ht]
		\centering
		\includegraphics[width=1.0\linewidth]{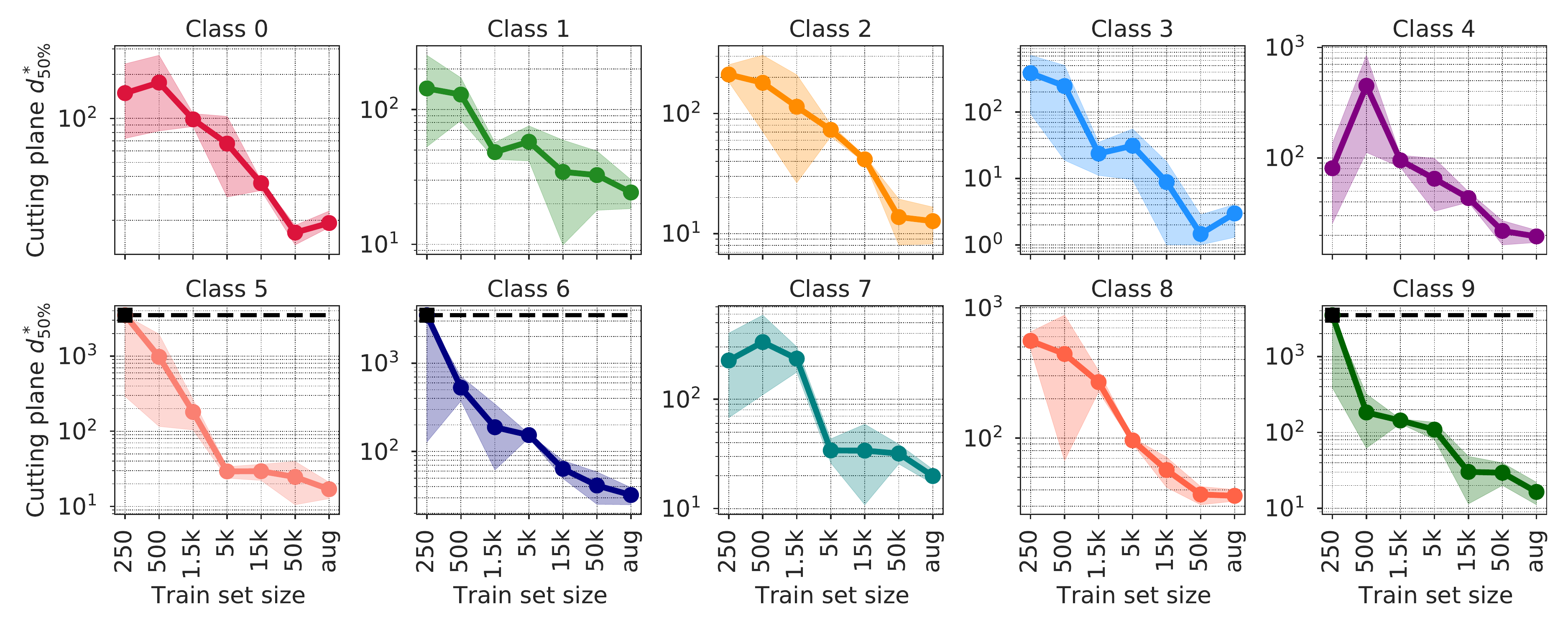}
		\caption{Comparison of the cutting plane dimension needed to get $50\%$ of the target class for ResNets trained to $100\%$ training accuracy on subsets of the training set of CIFAR-10 (mean and standard deviation of 2 networks shown). The bigger the training set, the smaller the $d^*_{\mathrm{50\%}}$, therefore the higher the class manifold dimension. The trend continues with the addition of data augmentation (aug), and takes the manifolds from low-D ($\ll D$) for small sets, to high-D ($\approx D$) for large set + augmentation. All classes shown in Fig.~\ref{fig:train_set_size_effect_all_classes}.}
		\label{fig:train_set_size_effect}
	\end{figure}

	During the course of training, a neural network has to learn to partition the $D$-dimensional space of inputs into generalizable regions of high class confidence that contain both the training points (directly enforced by loss minimization) and the test points (generalization). To see the role of training set size, we repeated our cutting plane experiments for networks trained to $100\%$ training set accuracy on subsets of CIFAR-10 of size 250, 500, 1.5k, 5k, 15k, and 50k images (=full training set) and added a final point using data augmentation on top, effectively mimicking a larger dataset. The bigger the training set, the smaller the $d^*_{\mathrm{50\%}}$, and therefore the larger the dimension of the CMs, as shown in Fig. \ref{fig:train_set_size_effect}. This trend held across all classes, and continued with augmentation. Better generalization is associated with higher CM dimensionality here.
	We hypothesize that the larger number of training points might allow the learned partitioning of the input space to connect previously disconnected and lower dimensional CMs through interpolation, thereby effectively increasing CM dimensions with training set size.
	
	\subsection{The effect of robustness to data corruptions.}
\begin{figure}[ht]
		\centering 
		\includegraphics[width=1.0\linewidth]{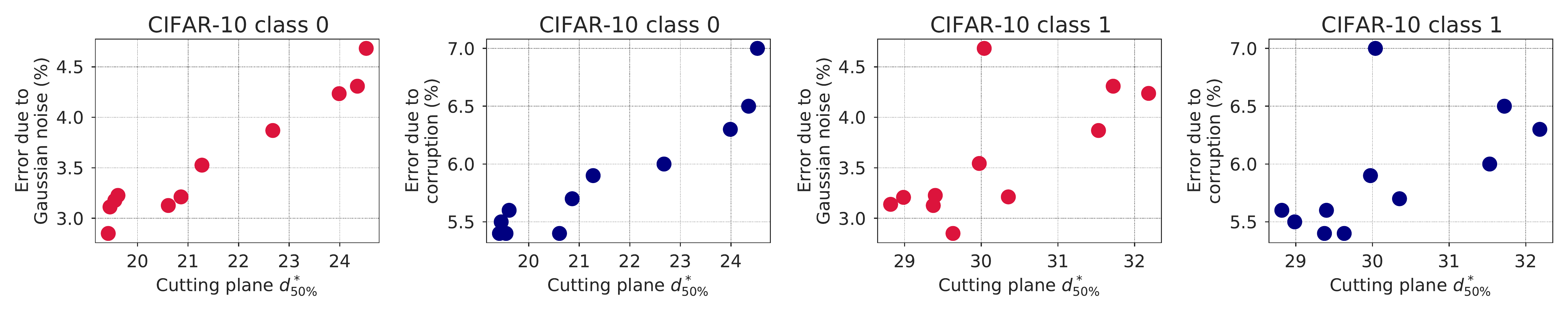}
		\caption{The correlation between class manifold dimension and model robustness to test-time distortions. The left panels show the error due to Gaussian noise applied at test time vs. $d^*_{\mathrm{50\%}}$ for classes 0 and 1 (restricted to due the high cost of the experiment). The right panels show the effect of $d^*_{\mathrm{50\%}}$ on error due to corruptions in CIFAR-10-C. Models with higher CM dimension are more robust to both Gaussian noise and to distortions in the Common Corruptions Benchmark. The plots show averages over a large number of models and random hyperparameter choices.}
		\label{fig:corruptions_effect}
	\end{figure}

	We measure the effect of cutting plane dimension on out-of-domain robustness of neural networks, which has recently been gaining in theoretical and practical importance (see e.g. \citet{ovadia2019trust}). Robustness to Gaussian noise was found to be a useful predictor for general robustness as well as adversarial robustness~\citep{ford2019adversarial,yin2019fourier}. For this reason, we first measure the robustness of WideResNet models to Gaussian noise applied at test time, where noise is sampled from a Gaussian with a standard deviation of 0.05, for each pixel independently. The left panel of Fig.~\ref{fig:corruptions_effect} shows the correlation between $d^*_{\mathrm{50\%}}$ and error due to Gaussian noise, calculated as the accuracy on corrupted data minus the accuracy on clean data. We see that the models with smaller $d^*_{\mathrm{50\%}}$, therefore higher class manifold dimension, are more robust to this type of noise.
	
	Next, we calculate the correlation between the $d^*_{\mathrm{50\%}}$ and the accuracy on CIFAR-10-C~\citep{hendrycks2018benchmarking}, which includes 15 different corruption types applied at test time (right panel of Fig.~\ref{fig:corruptions_effect}). These results together show that the effective dimension of neural networks class manifolds is correlated with their robustness to a variety of test-time distortions. The higher the class CM dimension, the better the robustness.                    
	Note that the results in Fig.~\ref{fig:corruptions_effect} are obtained across a large number of models and hyperparameter combinations to show the generality of the effect.
	
	\subsection{Evolution of dimension with training.}
\begin{figure}[ht]
		\centering
		\includegraphics[width=0.64\linewidth]{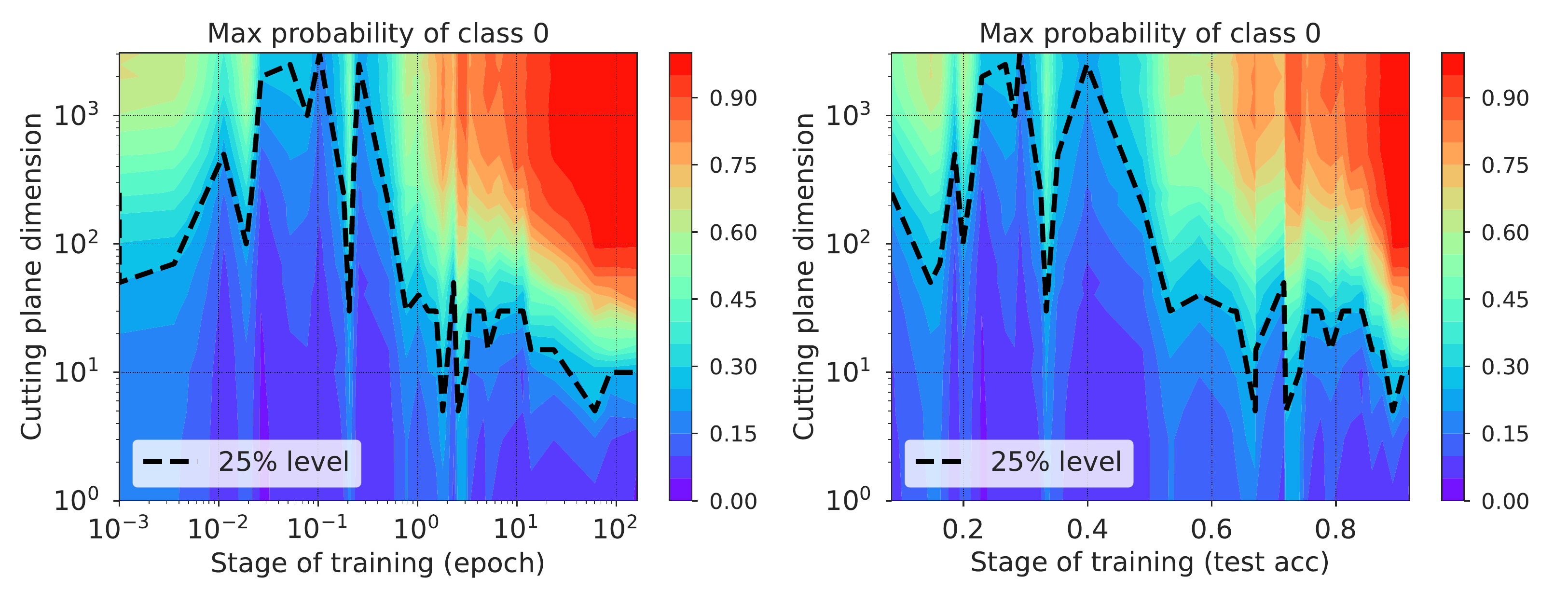}
		\includegraphics[width=0.34\linewidth]{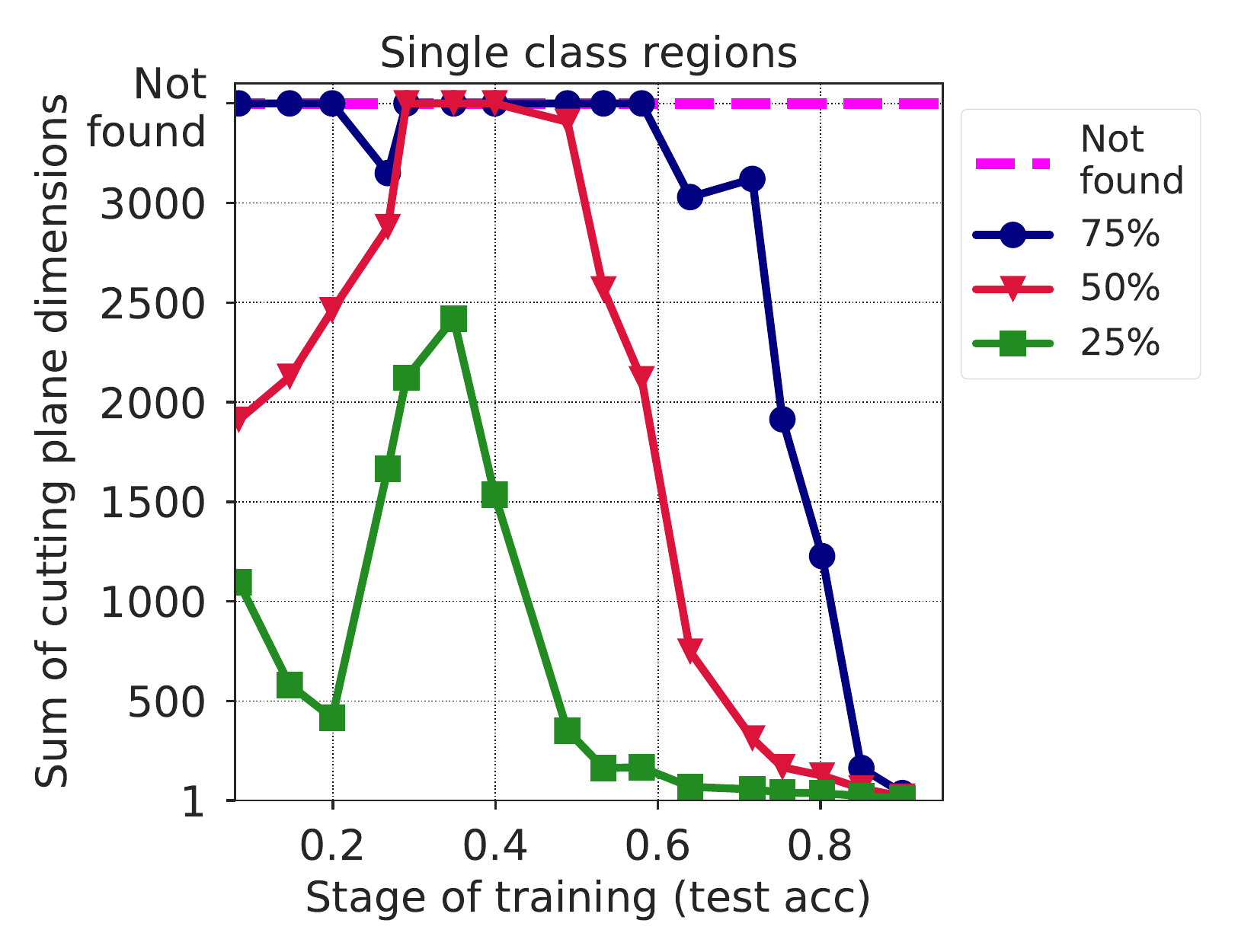}
		\caption{The effect of training on the dimension of cutting plane necessary to reach a particular probability. The two left panels show the maximum probability of class 0 reached for cutting planes of different dimensions over training for ResNet20v1 on CIFAR-10. The probability 25\% level (superset) is highlighted. 
			The right panel shows $d^*_{25\%}$, $d^*_{50\%}$ and $d^*_{75\%}$ for the average of all single class regions. 
			High confidence regions become hard to find at intermediate stages. Towards the end of training, the dimension of the manifolds grows (codimension $d^*$ goes down). The breakdown by classes is shown in Fig.~\ref{fig:single_class_dim_vs_test_acc_resnet_cifar10_2inits}.
		}
		\label{fig:effect_of_training}
	\end{figure}

	We study the effect of training on the high confidence manifolds in Fig.~\ref{fig:effect_of_training}. The early epochs are heavily influenced by the initialization. After a small amount of training, there seems to be an intermediate stage when it is very hard to find high confidence class manifolds ($d^*_{25\%}$ is high, and therefore the manifold dimension low). Towards the end of training, $d^*_{25\%}$ goes down for all classes (details in Figs.~ \ref{fig:single_class_dim_vs_test_acc_resnet_cifar10_2inits}, \ref{fig:single_class_dim_vs_test_acc_resnet_cifar100} and \ref{fig:single_class_dim_vs_test_acc_simplecnn_cifar10}). The non-monotonic behavior of the dimension points towards something unusual happening in the intermediate stages of training, and it could be related to the host of phenomena pointing towards the high impact of early stages of training. The causal reason for this remains for future work.
	
	\subsection{The effect of network width.}
\begin{figure}[ht]
		\centering
		\includegraphics[width=0.65\linewidth]{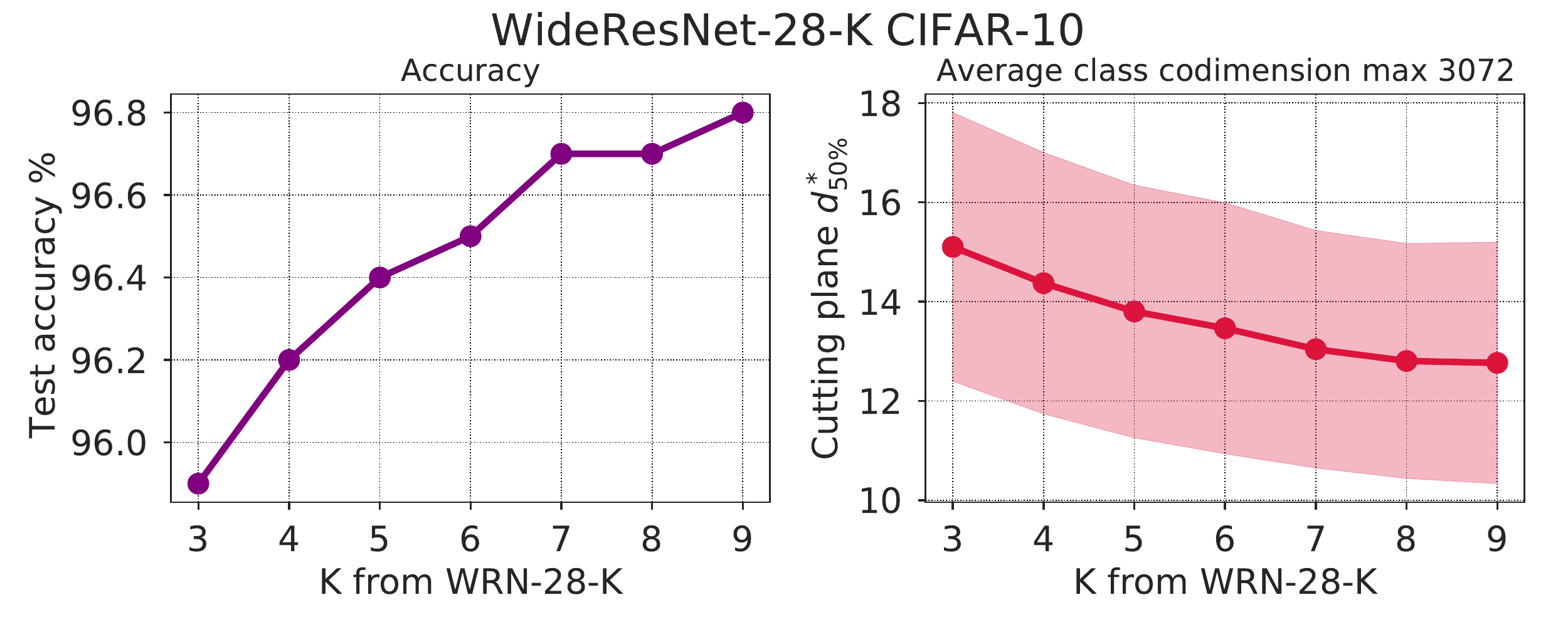}
		\caption{The effect of network width on the dimension. The left panel shows the final test accuracy of a WRN-28-K on CIFAR-10 for different values of the width $K$. The right panel shows $d^*_{50\%}$, the dimension of a cutting plane need to reach $50\%$ averaged over 10 classes (individual results shown in Fig.~\ref{fig:effect_of_width_all_classes}). The wider the network, the higher the dimension of the class manifolds.
		}
		\label{fig:effect_of_width_average_class}
	\end{figure}

	We found that the wider the neural network, the higher the dimension of the CMs (the smaller the dimension $d^*_{50\%}$). Our results for WideResNet-28-K \citep{zagoruyko2017wide}, where $K$ is specifying the width of the layers, are shown in Fig.~\ref{fig:effect_of_width_average_class} for the average of all CIFAR-10 classes (individual classes are in Fig.~\ref{fig:effect_of_width_all_classes}).  
	%
	
	\subsection{Model ensembling.}
\begin{figure}[!ht]
		\centering
		\includegraphics[width=0.34\linewidth]{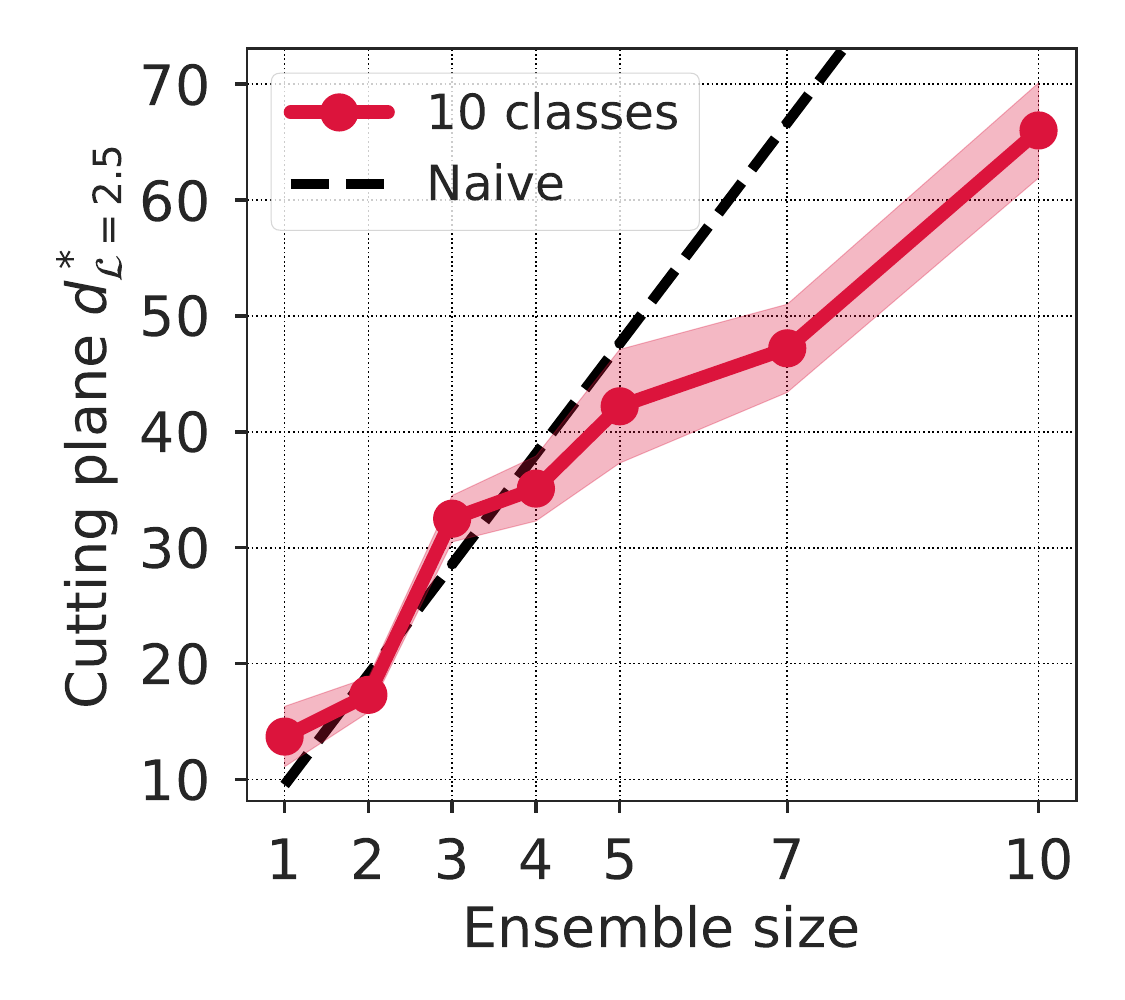}
		\includegraphics[width=0.58\linewidth]{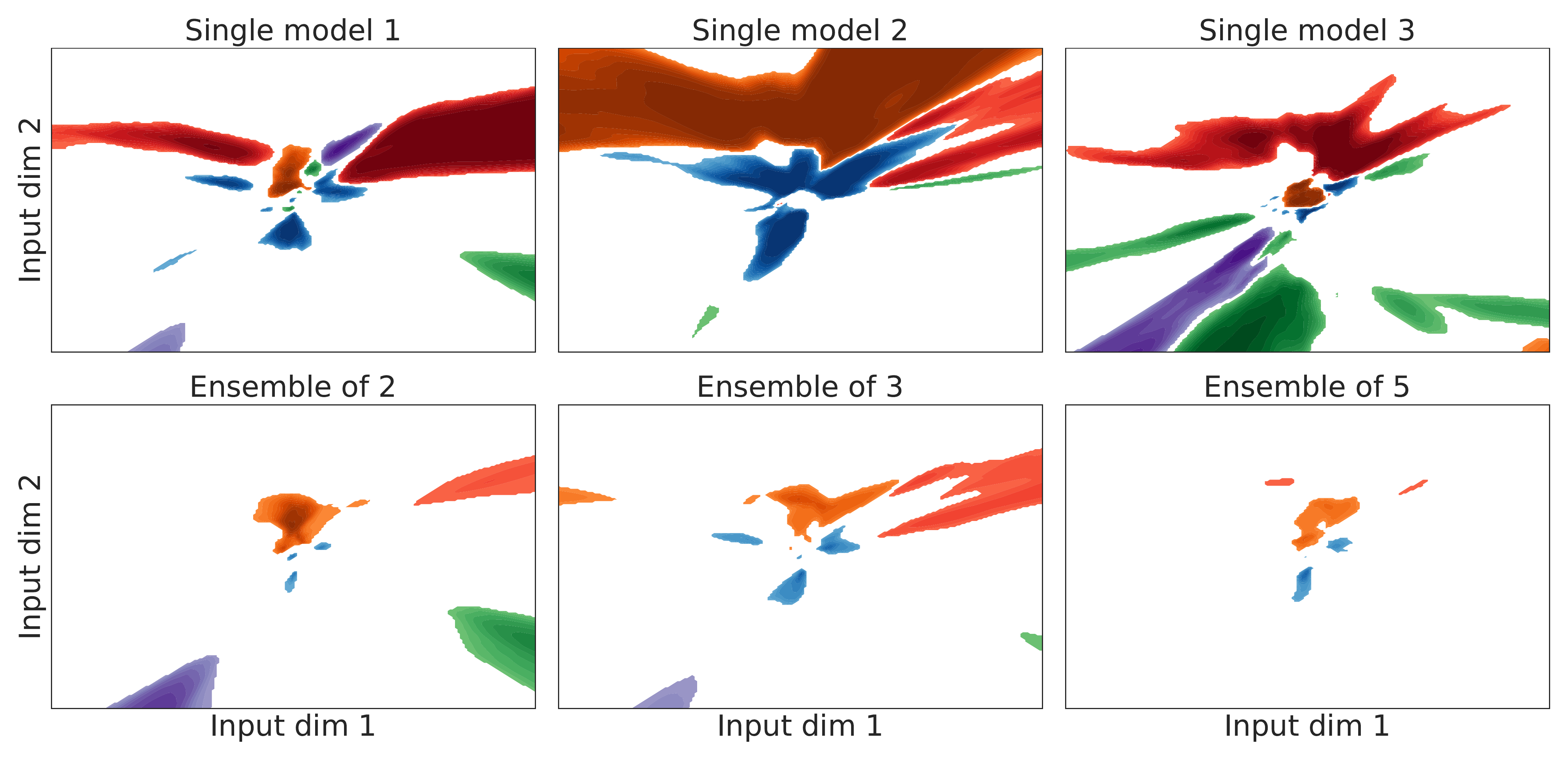}
		\includegraphics[width=0.78\linewidth]{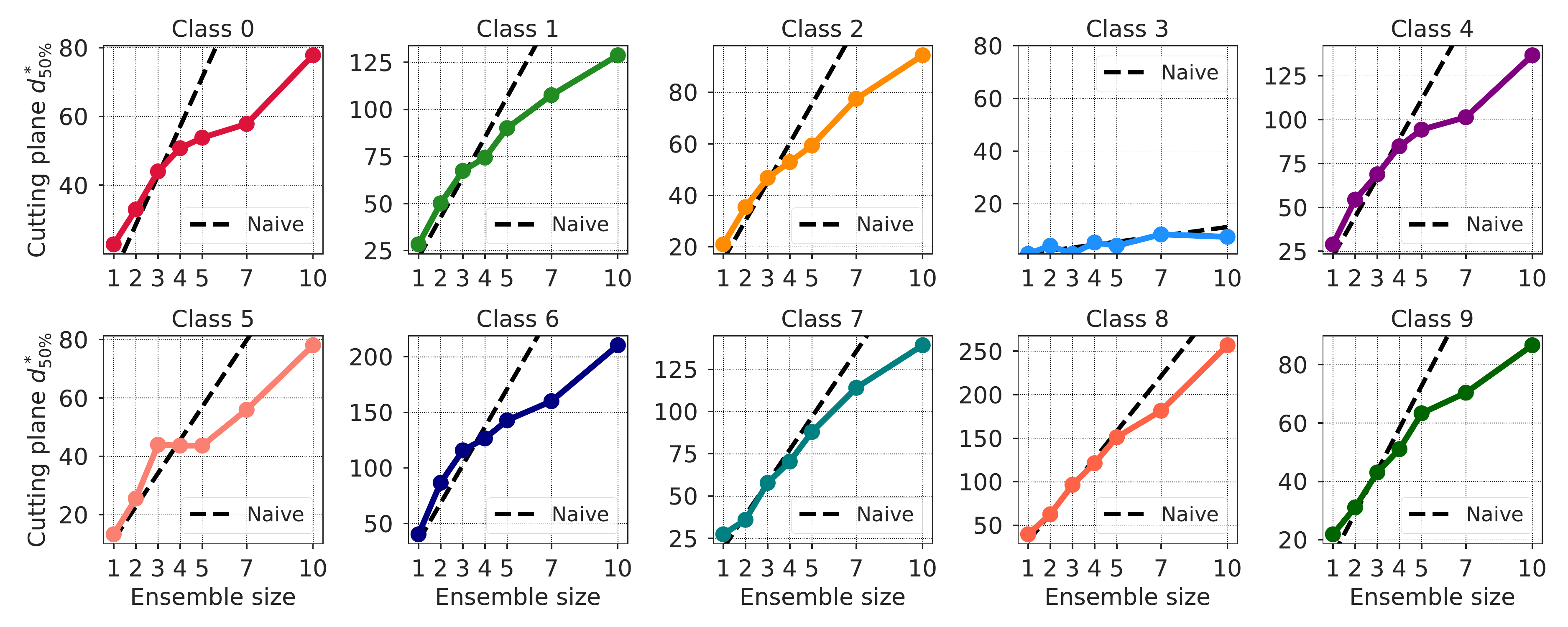}
		\caption{The effect of model ensembling on the dimension $d^*_{50\%}$  needed to reach $50\%$ accuracy averaged over all CIFAR-10 classes (top left) (individual class in the bottom row) for ResNet20 trained for 50 epochs. Across all classes, the larger the ensemble, the higher the  $d^*_{50\%}$ and therefore the lower the class manifold dimension. A naive model of addition of codimensions between models is overlayed, showing a surprisingly good fit for small ensembles. The right panels show a fixed random section of the input space for 3 single models (top row) and 3 ensemble sizes (bottom). The colors indicate 4 different classes $>50\%$. The elongated high-probability structures disappear with ensembled, as they get averaged.}
		\label{fig:resnet_ensemble}
		\label{fig:ensemble_by_class}
	\end{figure}

	We found that model ensembling (taking $N$ independently trained models, giving them the same input, and averaging their predicted probabilities) reliably leads to class manifolds of lower dimension, as well as between-class regions of lower dimension. The bigger the ensemble, the lower the dimension, as shown in a summary plot in Fig.~\ref{fig:resnet_ensemble} (average over all 10 classes). This is atypical, as all other methods of improving performance (e.g. larger training set, more training (towards the end), width) correlated with higher dimensional CMs. This suggests that ensembles might be doing something geometrically distinct from the other methods. This could be related to the observation that, unlike other techniques, deep ensembles combine models from distinct loss landscape basins \cite{fort2020deep}, which can be partially reached by architectures such as MIMO \cite{havasi2020training}.
	In Fig.~\ref{fig:ensemble_by_class} we show the effect of ensemble size on the class manifold codimension individually for all 10 classes of CIFAR-10. A very simple model, predicting that the codimension of class manifold for an ensemble of $N$ models scales linearly with $N$ is born out well for small $N$ there. This is supported by the right panels in Fig.~\ref{fig:resnet_ensemble} which show a 2D section of the input space with class regions of different classes highlighted for 3 different models and ensembles of different sizes. The class regions seem relatively randomly oriented, leading to the addition of manifold codimensions, explaining the relation $\mathrm{codim} \propto N$ observed in Fig.~\ref{fig:ensemble_by_class}.
	%
	%
	\section{Conclusion}
	We propose a new tool that we call the Subspace Tomography method for estimating the dimension of class manifolds and multi-way class boundary manifolds in the space of inputs for deep neural networks. To circumvent the curse of dimensionality, we use optimization constrained to randomly chosen affine subspaces (cutting planes) of varying dimension. This allows us to extract the effective dimension of the class manifolds as well regions between classes. Our mathematical analysis uses the concept of Gaussian width and the Gordon's escape through mesh theorem from high-dimensional geometry to define a robust, effective dimension. We study the manifold dimension as a function the network, architecture, stage of training, accuracy and robustness and find a ubiquitous correlation between higher class manifold dimension and better performance and robustness along the many axes tested points towards an intimate link between the geometry of the input space class partitioning and generalization. Ensembling is the only technique amongst the ones we explored that both increases performance and decreases the manifold dimension at the same time, suggesting that its beneficial effects might be geometrically distinct from other ways of improving performance.
	
	 \subsubsection*{Acknowledgments}
	We would like to thank Ilya Tolstikhin from Google Research Zurich who was instrumental in the early phases of development of this project, and Dustin Mixon from Ohio State University for discussions on Gordon's escape through a mesh theorem.

	\clearpage
	\bibliography{input_cuts_arxiv.bib}
	\bibliographystyle{unsrtnat}
	
	\clearpage
	
	\appendix
	\section{Appendix}
	
	\subsection{Details of networks, datasets and training}
	\label{sec:training_details}
	In this paper we use two architectures: 1) \verb|SimpleCNN|, which is a simple 4-layer CNN with 32, 64, 64 and 128 channels, $\mathrm{ReLU}$ activations and \verb|maxpool| after each convolution, followed by a fully-connected layer, and 2) \verb|ResNet20v1| as described in \citet{he2015deep} with batch normalization on \citep{IofSze15}. We also analyze the Vision Transformer \cite{dosovitskiy2020image} pretrained on ImageNet and finetuned on CIFAR-10. We use 5 datasets: MNIST \citep{lecun-mnisthandwrittendigit-2010}, Fashion MNIST \citep{xiao2017fashionmnist}, CIFAR-10 and CIFAR-100 \citep{cifar10}, and ImageNet \cite{imagenet_cvpr09}. The ResNet is trained for 200 epochs using SGD+Momentum at learning rate $0.1$, dropping to $0.01$ at epoch 80 and $0.001$ at epoch 120. The $L_2$ norm regularization is $10^{-4}$. In one experiment, we use data augmentation as described in \footnote{\url{https://github.com/keras-team/keras/blob/master/examples/cifar10_resnet.py}}. For our robustness experiments, we used the Wide-ResNet models~\citep{zagoruyko2016wide} available in \footnote{\url{https://github.com/tensorflow/models/tree/master/research/autoaugment}}. We trained 11 different sizes of Wide-ResNet models (WRN-28-2 to WRN-28-12) with AutoAugment. Each model was trained from 15 different random weight-initializations for better statistics. We used the following hyperparameters to train each model: a learning decay of 0.1, weight decay of 5e-4, cosine learning rate decay in 200 epochs, and AutoAugment~\citep{cubuk2018autoaugment} for data augmentation. 
	
	To see what the effect of diverse architectures were on our conclusions,  we experimented with the new Vision Transformer \citep{dosovitskiy2020image} that was pretrained on ImageNet and finetuned on CIFAR-10, as recommended in their published code\footnote{\url{https://github.com/google-research/vision_transformer}}.
	
	\subsection{Detailed derivation of the closest approach of two affine subspaces}s
	\label{sec:analytics}
	\label{sec:derivation}
	Let us consider a situation in which in a $D$-dimensional space we have a randomly chosen $d$-dimensional affine subspace $A$ defined by a point $\vec{X}_0 \in \mathbb{R}^D$ and a set of $d$ orthonormal basis vectors $\{ \hat{v}_i \}_{i = 1}^{d}$ that we encapsulate into a matrix $M \in \mathbb{R}^{d \times D}$. Let us consider another random $n$-dimensional affine subspace $B$. Our task is to find a point $\vec{X}^* \in A$ that has the minimum $L_2$ distance to the subspace $B$, mathematically $\vec{X}^* = \mathrm{argmin}_{\vec{X} \in A} \left | \vec{X} -  \mathrm{argmin}_{\vec{X^\prime} \in B} \left | \vec{X} - \vec{X}^\prime \right | \right | $. In words, we are looking for a point in the $d$-dimensional subspace $A$ that is as close as possible to its closest point in the $n$-dimensional subspace $B$. A point within the subspace $A$ is parametrized by a $d$-dimensional vector $\vec{\theta} \in \mathbb{R}^d$ by $\vec{X}(\theta) = \vec{\theta} M + \vec{X}_0 \in A$. This parametrization ensures that for all choices of $\vec{\theta}$ the resulting $\vec{X} \in A$.   
	
	Without loss of generality, let us consider the case where the $n$ basis vectors of the subspace $B$ are aligned with the dimensions $D-n,D-n+1,\dots,D$ of the coordinate system. Let us call the remaining axes $s = D-n$ the \textit{short} directions of the subspace $B$. A distance from a point $\vec{X}$ to the subspace $B$ now depends only on its coordinates $1,2,\dots,s$. Therefore $l^2(\vec{X},B) = \sum_{i=1}^s X_i^2$. This is the case because of our purposeful choice of coordinates.
	
	Given that the only coordinates influencing the distance are the first $s$ values, let us, without loss of generality, consider a $\mathbb{R}^s$ subspace of the original $\mathbb{R}^D$ only including those. Then the distance between a point within the subspace $A$ parametrized by the vector $\vec{\theta}$ is $l^2(\vec{X}(\theta),B) = \left | \vec{\theta} M + \vec{X}_0 \right |^2$. Given our restrictions, now the $\theta \in \mathbb{R}^d$, $M \in \mathbb{R}^{d \times s}$ and $\vec{X}_0 \in \mathbb{R}^d$. The distance $l$ attains its minimum for $\partial_{\vec{\theta}} l^2 = \left ( \vec{\theta} M + \vec{X}_0 \right ) M^T = \vec{0}$, producing the minimality condition $\vec{\theta}^* M = -\vec{X_0}$. There are now 3 cases:
	
	\textbf{1. The overdetermined case, $d > s$.} In case $d > s = D-n$, the optimal $\theta^* = - \vec{X}_0 M^{-1}$ belongs to a ($d-s = d+n-D$)-dimensional family of solutions that attain $0$ distance to the plane $B$. In this case the affine subspaces $A$ and $B$ intersect and share a ($d+n-D$)-dimensional intersection.
	
	\textbf{2. A unique solution case, $d = s$.} In case of $d = s = D-n$, the solution is a unique $\theta^* = - \vec{X}_0 M^{-1}$. After plugging this back to the distance equation, we obtain $\vec{\theta}$ is $l^2(\vec{X}(\vec{\theta}^*),B) = \left | - \vec{X}_0 M^{-1} M + \vec{X}_0 \right |^2 = \left | - \vec{X}_0 + \vec{X}_0 \right |^2 = 0$. The square (in this case) matrix $M$ and its inverse $M^{-1}$ cancel each other out. 
	
	\textbf{3. An underdetermined case, $d < s$.} In case of $d < s$, there is generically no intersection between the subspaces. The inverse of $M$ is now the Moore-Penrose inverse $M^+$. Therefore the closest distance is $\vec{\theta}$ is $l^2(\vec{X}(\vec{\theta}^*),B) = \left | - \vec{X}_0 M^{+} M + \vec{X}_0 \right |^2$. Before our restriction from $D \to s$ dimensions, the matrix $M$ consisted of $d$ $D$-dimensional, mutually orthogonal vectors of unit length each. We will consider these vectors to be component-wise random, each component with variance $1/\sqrt{D}$ to satisfy this condition on average. After restricting our space to $s$ dimensions, $M$'s vectors got reduced to $s$ components each, keeping their variance $1/\sqrt{D}$. They are still, in expectation, mutually orthogonal, however, their length got reduced to $\sqrt{s} / \sqrt{D}$. The (transpose) of the inverse $M^+$ consists of vectors of the same directions, with their lengths scaled up to $\sqrt{D} / \sqrt{s}$. That means that, in expectation, $M M^+$ is a diagonal matrix with $d$ diagonal components set to $1$, and the remainder being $0$. The matrix $(I - M^+ M)$ contains $(s-d)$ ones on its diagonal. The projection $|\vec{X}_0 (I - M^+ M)|^2$ is therefore of the expected value of $|X_0|^2 (s-d)^2 / D$. The expected distance between the $d$-dimensional subspace $A$ and the $d$-dimensional subspace $B$ is, in expectation
	\begin{equation}
	\mathbb{E}\left [ d(A,B) \right ] \propto \begin{cases} \frac{\sqrt{D - n - d}}{\sqrt{D}} & n + d < D \, , \\ 0 & n + d \ge D \, . \end{cases} 
	\label{eqn:distance_analytic_scaling}
	\end{equation}
	We ran a numerical experiment using automatic differentiation in JAX \citep{jax2018github} where we generated random affine subspaces of different dimensions and measured their closest approach using optimization to locate the place. The numerical results presented in Figure \ref{fig:toy_model_comparison} match the analytic predictions in Equation \ref{eqn:distance_analytic_scaling} well.
	
	\subsection{Empirical fit function}
	The empirical fit function we use to extract the critical dimension of the cutting hyperplane $d^*_{50\%}$ is shown in \ref{eq:fit}. 
	\begin{equation}
	p(d; A,B,C,D) = A + \frac{B}{1 + \exp \left ( -\log ( d / C) / D \right )} \, .
	\end{equation}
	It is a sigmoid function that depends logarithmically on the dimension $d$ and can be offset from $p=0$ at for low $d$ and from $p=1$ for high $d$. That is the case as sometimes the neural networks we analyzed would not have any regions of a particular class reaching all the way to $100\%$. In other cases, even optimization in a line $d=1$ would be able to get to a $p>10\%$ (for 10 class classification).
	
	For fitting the loss $\mathcal{L}(d)$, we utilized the fact that the cross-entropy loss depends logarithmically on $p$, and therefore used
	\begin{equation}
	\mathcal{L}(d; A,B,C,D) = -\log \left [ A + \frac{B}{1 + \exp \left ( -\log ( d / C) / D \right )} \right ] \, .
	\label{eq:fit}
	\end{equation}
	In both cases $A$, $B$, $C$ and $D$ are free fit parameters. We used SciPy optimizer \citep{2020SciPy-NMeth} to find the parameters and their covariance.  
	
	\subsection{Cutting plane axis-alignment -- the effect of sparsity}
	When choosing the matrix $M$ that defines the span of the subspace in which we are optimizing, we can choose to make the rows of $M$ sparse. On one end, each basis vectors might generically be non-zero in each of its components, while on the other end, a single non-zero element per basis vector is allowed. 
	\begin{figure}[ht]
		\centering
		\includegraphics[width=0.36\linewidth]{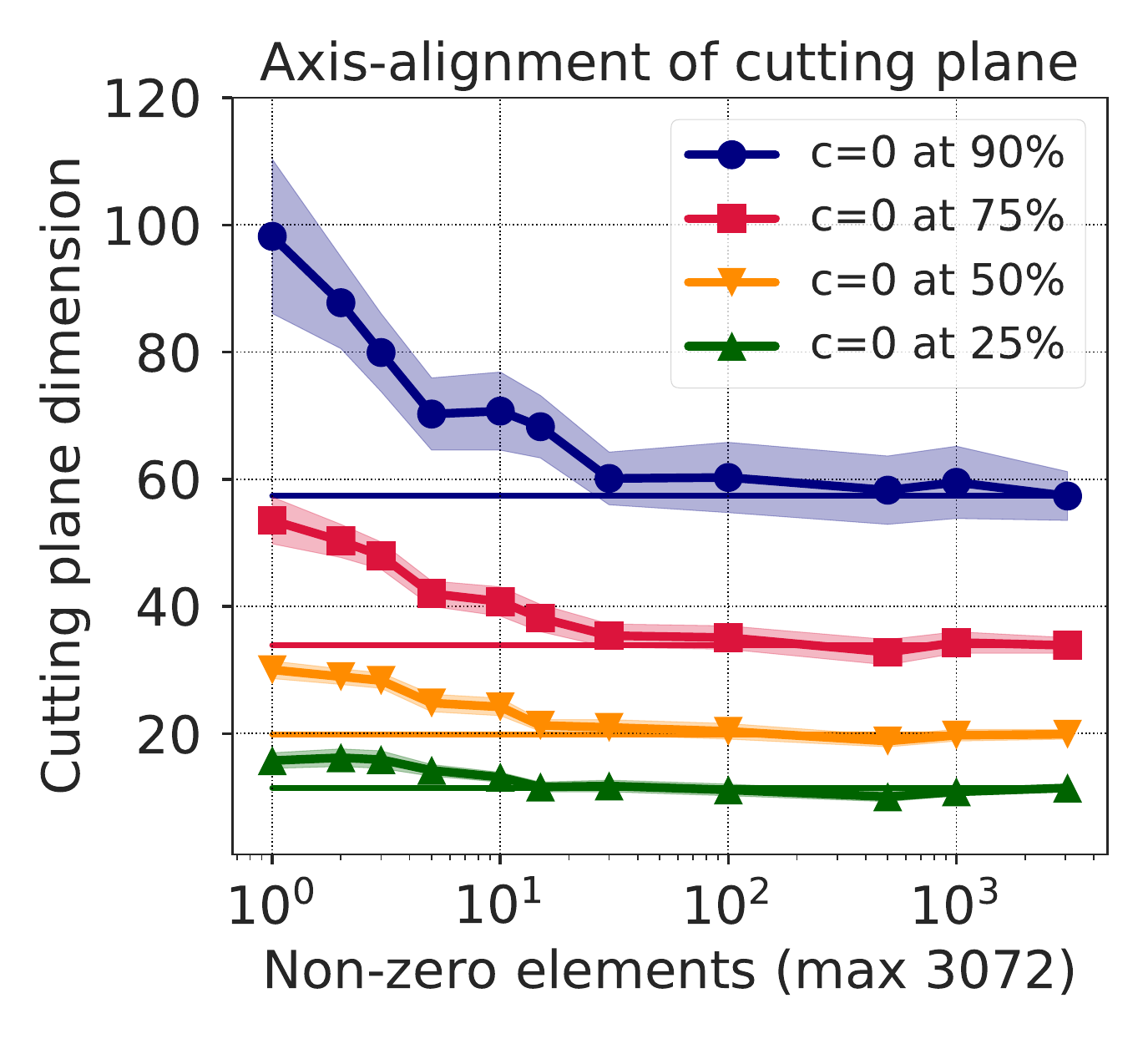}
		\raisebox{5mm}{\includegraphics[width=0.63\linewidth]{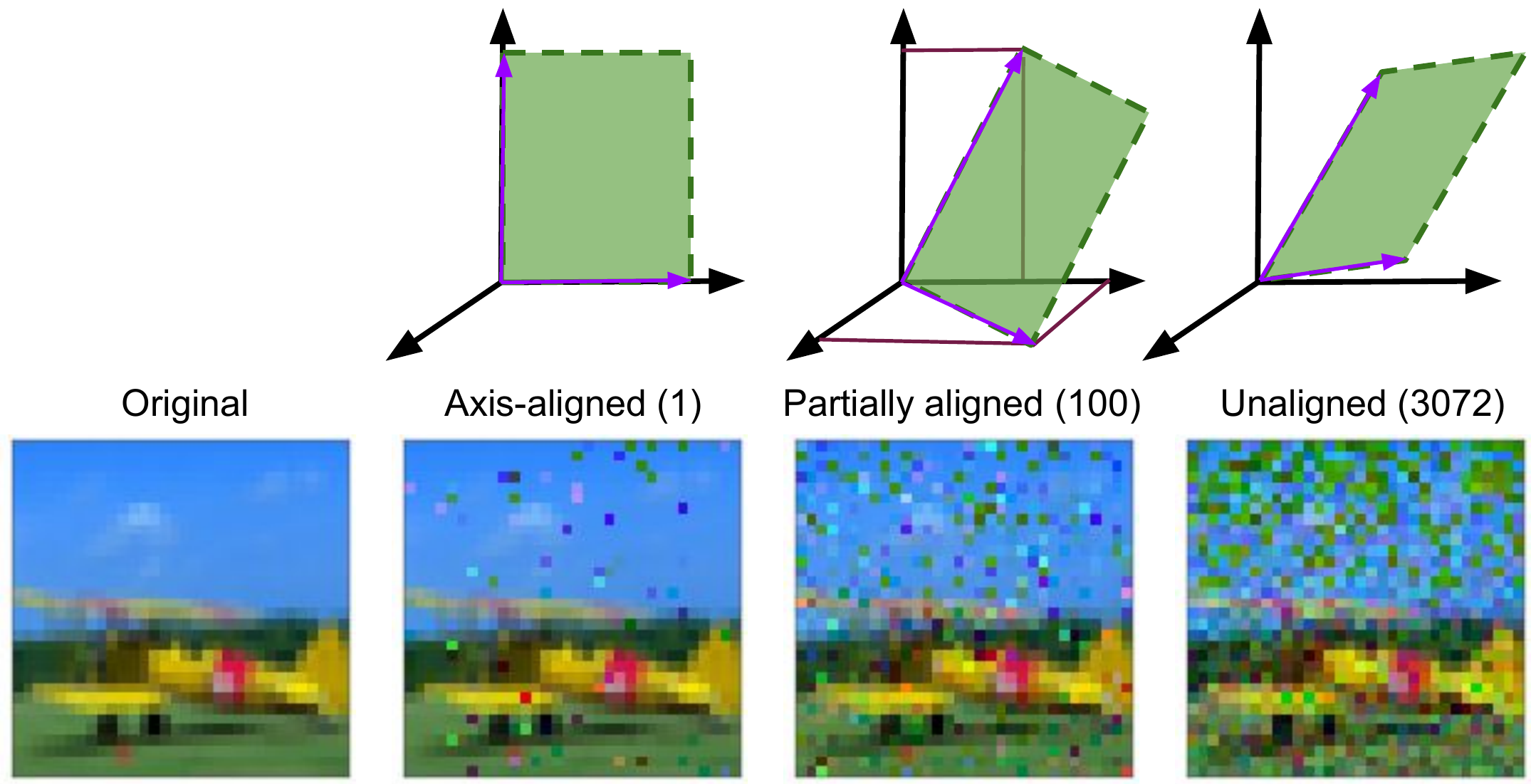}}
		\caption{The effect of axis alignment of the cutting planes. The figure shows the cutting plane dimension necessary to reach 4 thresholds levels (the 4 data lines) of class 0 probability (y-axis) from a random starting point for a well trained ResNet20v1 on CIFAR-10. We vary the number of non-zero elements of the basis vectors of the random cutting plane (x-axis). For a small number of non-zero elements, single pixels are varied, while for a 3072 non-zero elements (the maximum value), all pixels are varied jointly. The axis-aligned random cuts require higher dimensions to hit the same accuracy regions of class 0.}
		\label{fig:plane_alignemnt_effect}
	\end{figure}
	Geometrically, this corresponds to the alignment of the subspace with the axes (pixels and their channels for images) of the input space. Figure \ref{fig:plane_alignemnt_effect} shows the effect of the sparsity of $M$ on the resulting $d^*_{25\%}$, $d^*_{50\%}$, $d^*_{75\%}$ and $d^*_{90\%}$. The sparser the $M$, the higher the dimension needed to reliably reach the $25\%$, $50\%$, $75\%$, and $90\%$ class confidence region respectively. The effect of sparsity is visible, however, it is 1) not very significant (changing the dimension by a small part of the total $D=3072$ for CIFAR-10), and 2) its effect disappears for even small amounts of non-zero elements in $M$. 
	
	\subsection{Training on randomly permuted labels}
\begin{figure}[ht]
		\centering
		\includegraphics[width=1.0\linewidth]{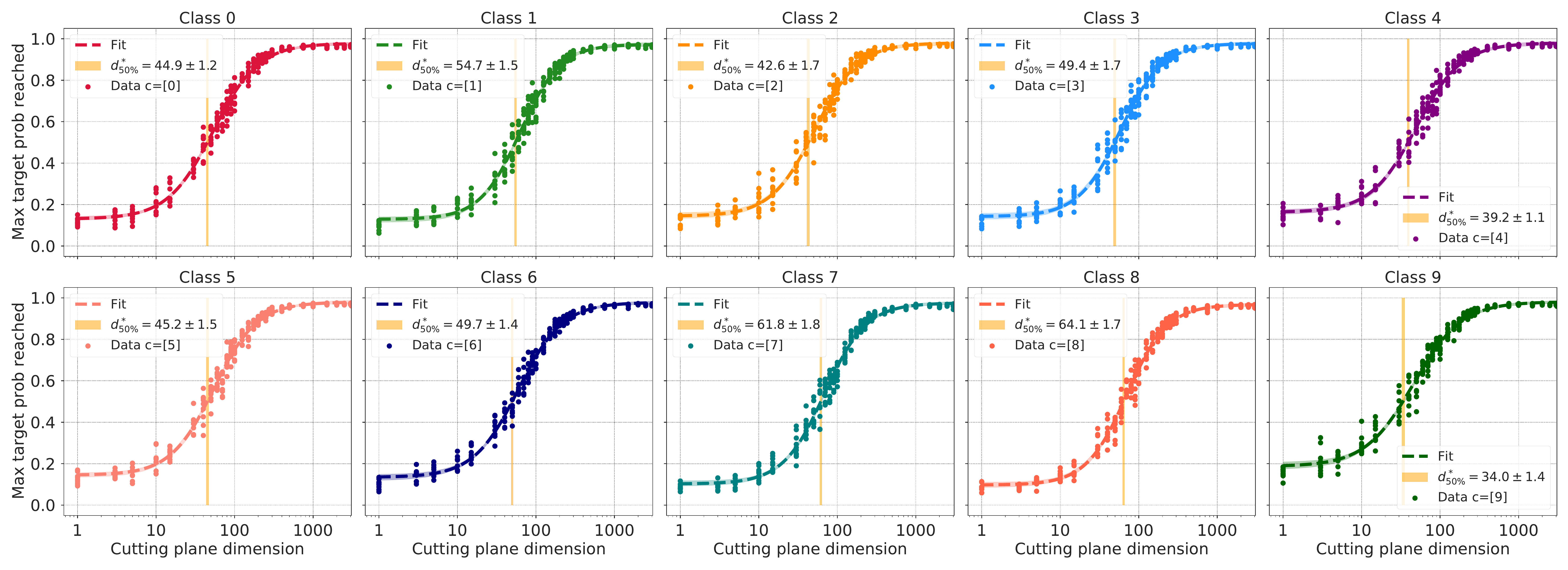}
		\includegraphics[width=1.0\linewidth]{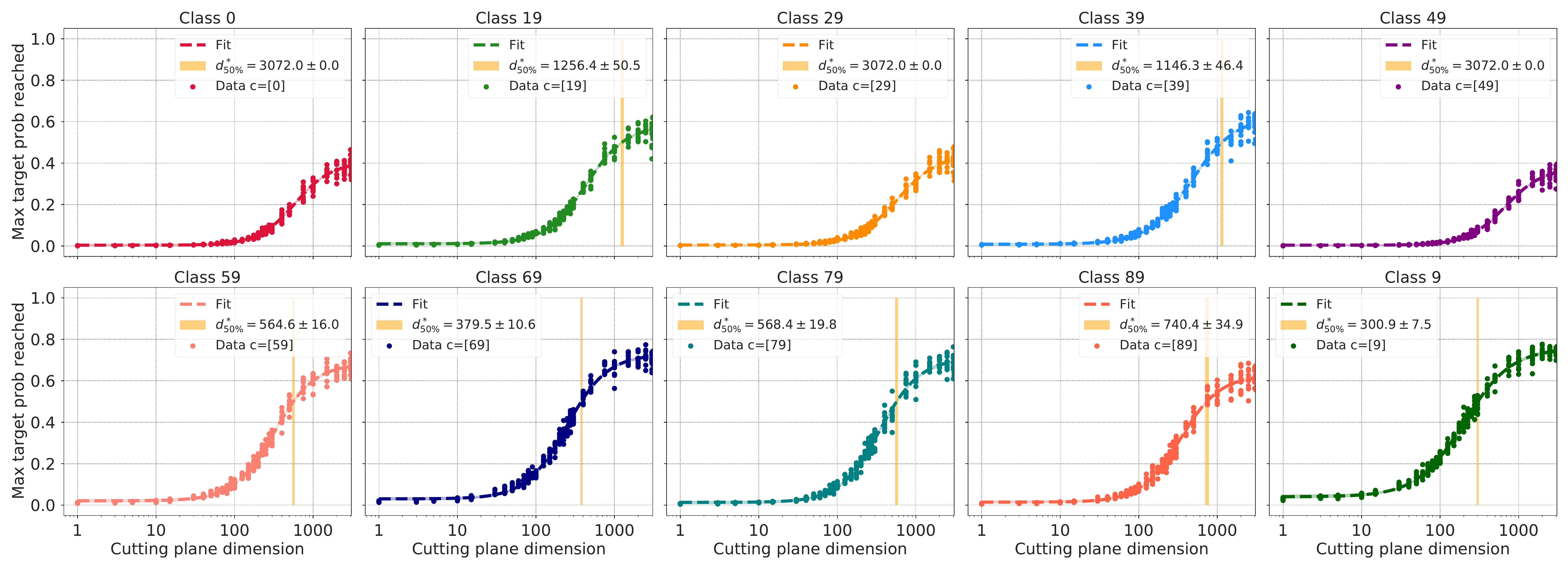}
		\caption{Maximum probability reached on cutting planes of different dimensions for all 10 target classes of CIFAR-10 (top row) and CIFAR-100 (bottom row) for a ResNet20v1 trained to 100\% training accuracy on \textit{randomly permuted} class labels. The $d^*_{\mathrm{50\%}}$ is consistently higher and therefore the dimension of the high confidence manifolds is lower than for semantically meaningful labels (Figure \ref{fig:single_class_prob_curves_cifar10}), suggesting geometrically a very different function being learned.}
		\label{fig:permuted_single_class_resnet_cifar10}
	\end{figure}

	For training on randomly permuted labels of the training set, we observe the critical dimension $d^*_{50\%}$ to rise significantly, meaning that a much higher dimensional cutting plane is needed to reliably intersect a class manifold. The breakdown by class for ResNet20v1 on CIFAR-10 and CIFAR-100 is shown in Figure \ref{fig:permuted_single_class_resnet_cifar10}. 
	The comparison to semantically meaningful labels is shown in Figure \ref{fig:models_plot}.

	\subsection{Additional cutting curves for CIFAR-10 and CIFAR-100}
\begin{figure}[ht]
		\centering
		\includegraphics[width=1.0\linewidth]{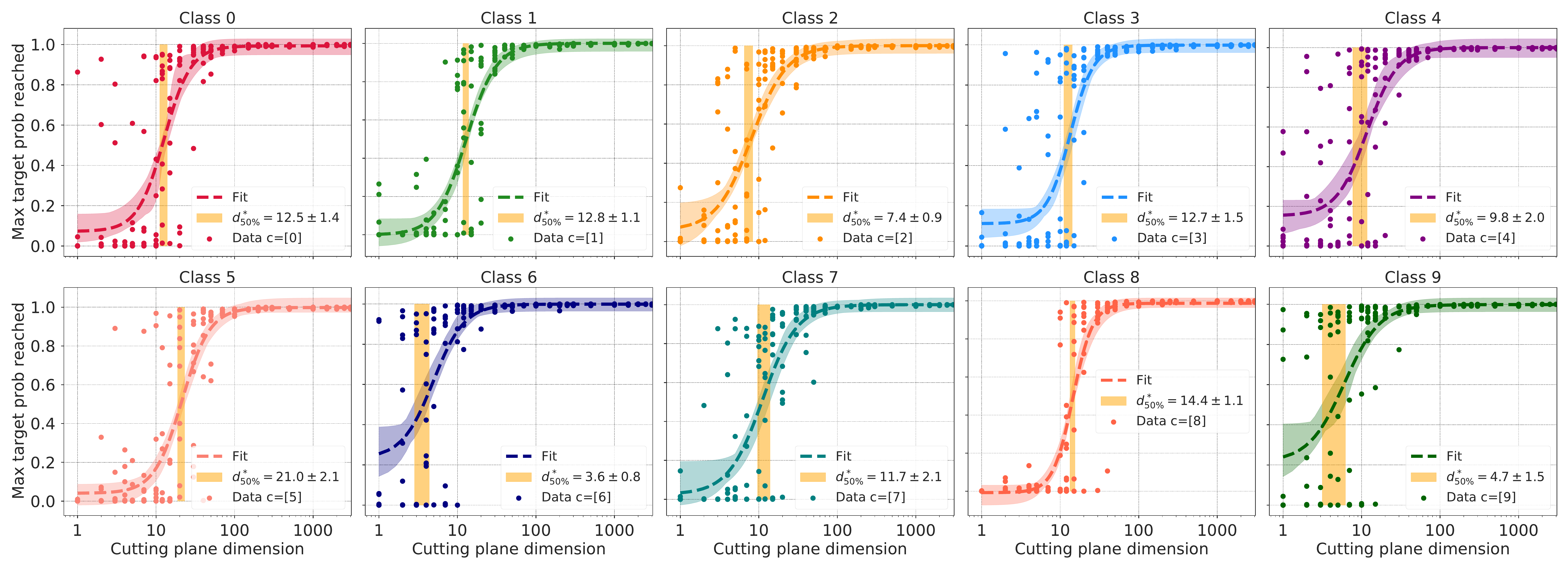}
		\caption{Maximum probability of single classes of CIFAR-10 reached on cutting planes of dimension $d$. The figure shows the dependence of the probability of a single class of CIFAR-10 (y-axes) reached on random cutting hyperplanes of different dimensions (x-axes). The results shown are for a well-trained ($>76\%$ test accuracy) SimpleCNN on CIFAR-10. Each dimension $d$ is repeated $10\times$ with random planes and offsets.}
		\label{fig:single_class_prob_curves_simpleCNN_cifar10}
	\end{figure}

\begin{figure}[ht]
		\centering
		\includegraphics[width=1.0\linewidth]{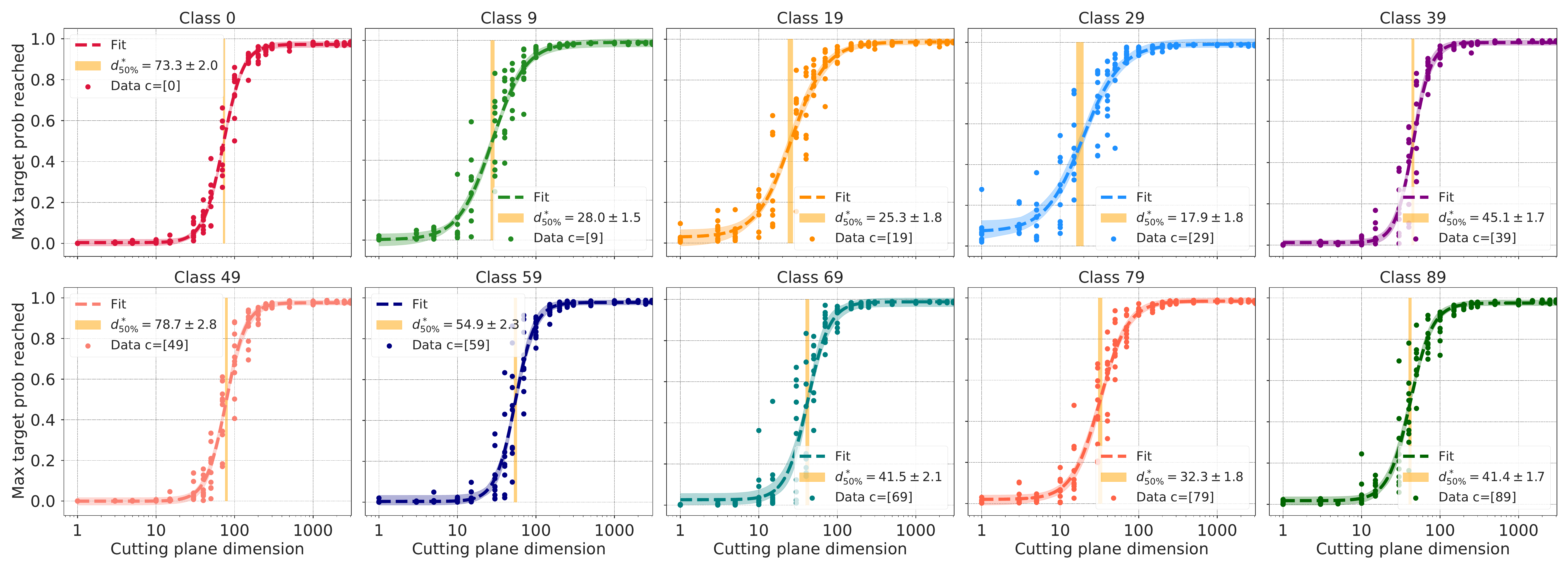}
		\caption{Maximum probability of selected single classes of CIFAR-100 reached on cutting planes of dimension $d$. The figure shows the dependence of the probability of a single class of CIFAR-100 (y-axes) reached on random cutting hyperplanes of different dimensions (x-axes). The results shown are for a well-trained ($>67\%$ test accuracy) ResNet20v1 on CIFAR-100. Each dimension $d$ is repeated $10\times$ with random planes and offsets.}
		\label{fig:single_class_prob_curves_resnet_cifar100}
	\end{figure}

	Two additional detailed cutting plane results can be found in this subsection: SimpleCNN on CIFAR-10 in Figure \ref{fig:single_class_prob_curves_simpleCNN_cifar10}, and ResNet20v1 on CIFAR-100 in Figure \ref{fig:single_class_prob_curves_resnet_cifar100}.
	\begin{figure}[ht]
		\centering
		\includegraphics[width=1.0\linewidth]{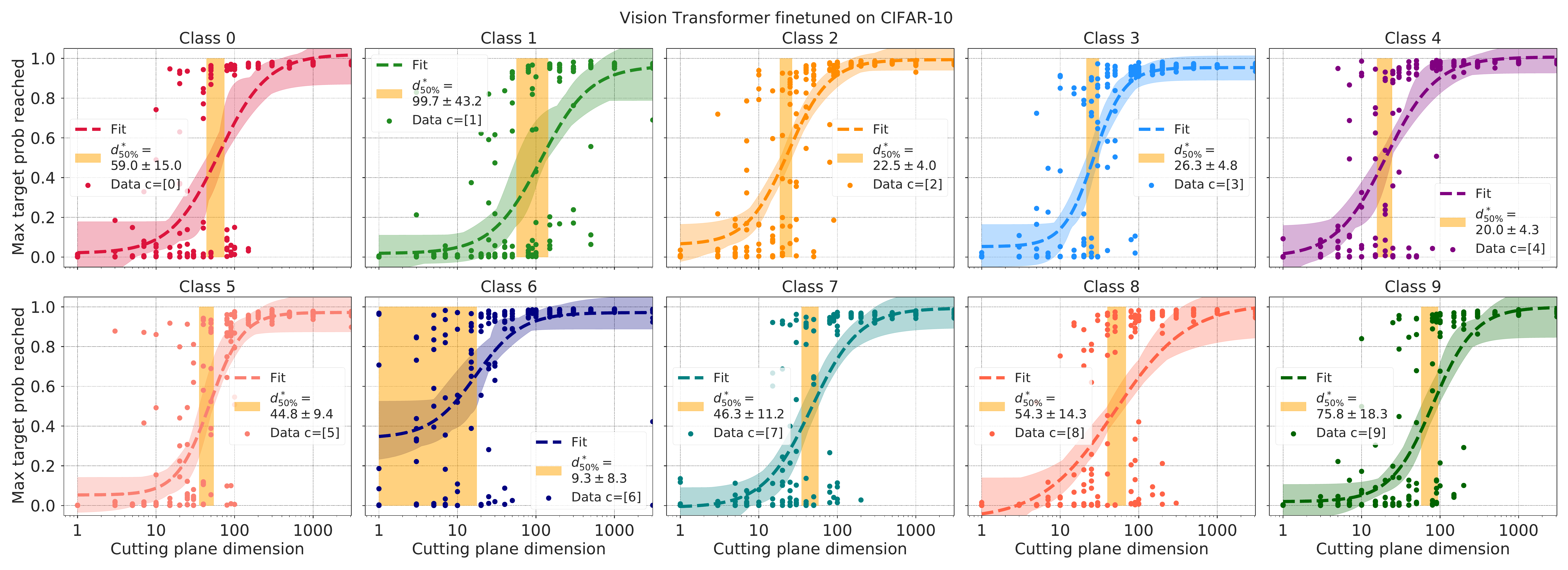}
		\caption{Maximum probability of all classes of CIFAR-10 reached on cutting planes of dimension $d$. The figure shows the dependence of the probability of a single class of CIFAR-10 (y-axes) reached on random cutting hyperplanes of different dimensions (x-axes). The results shown are for a well-trained Vision Transformer pre-trained on ImageNet and finetuned to CIFAR-10 to test accuracy $>97\%$. Each dimension $d$ is repeated $10\times$ with random planes and offsets.}
		\label{fig:single_class_prob_curves_vision_transformer_cifar10}
	\end{figure}
	
	\subsection{Dimension as a function of training stage}
\begin{figure}[ht]
		\centering
		\includegraphics[width=1.0\linewidth]{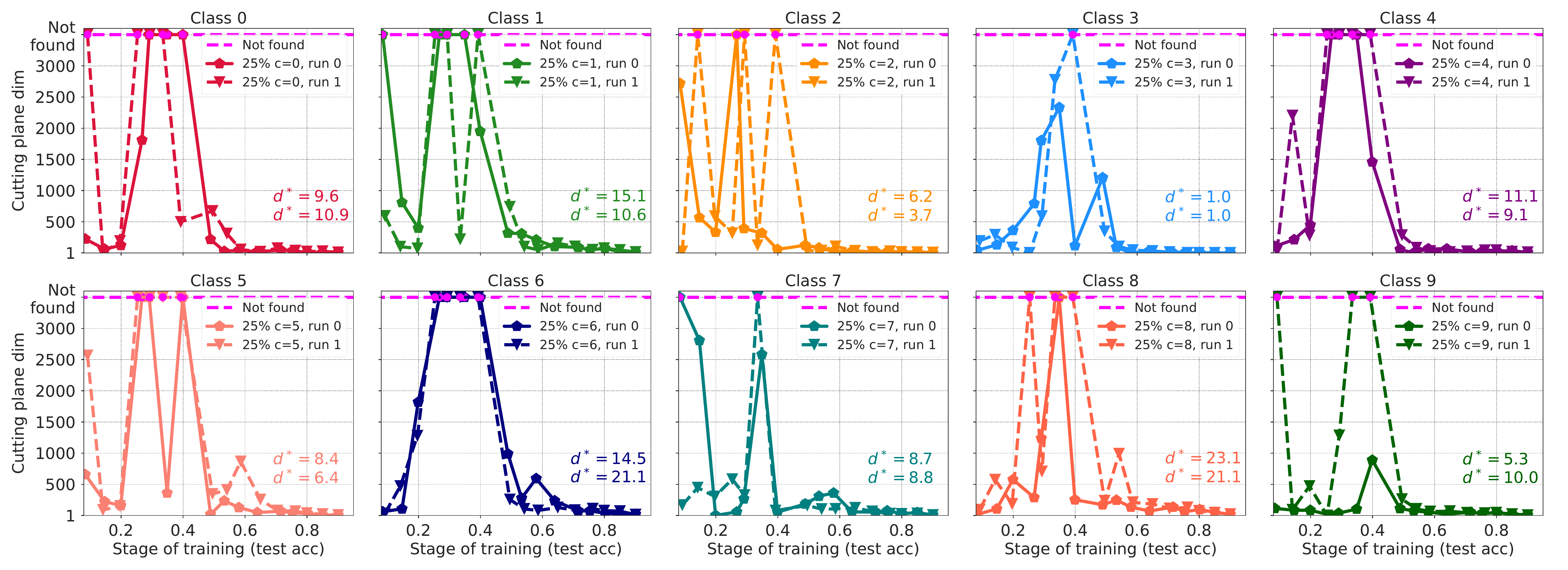}
		\caption{The cutting plane dimension needed to reach $25\%$ probability for the 10 classes of CIFAR-10 as a function of training stage for a ResNet20v1, averaged over two initializations and runs.}
		\label{fig:single_class_dim_vs_test_acc_resnet_cifar10_2inits}
	\end{figure}

\begin{figure}[ht]
		\centering
		\includegraphics[width=1.0\linewidth]{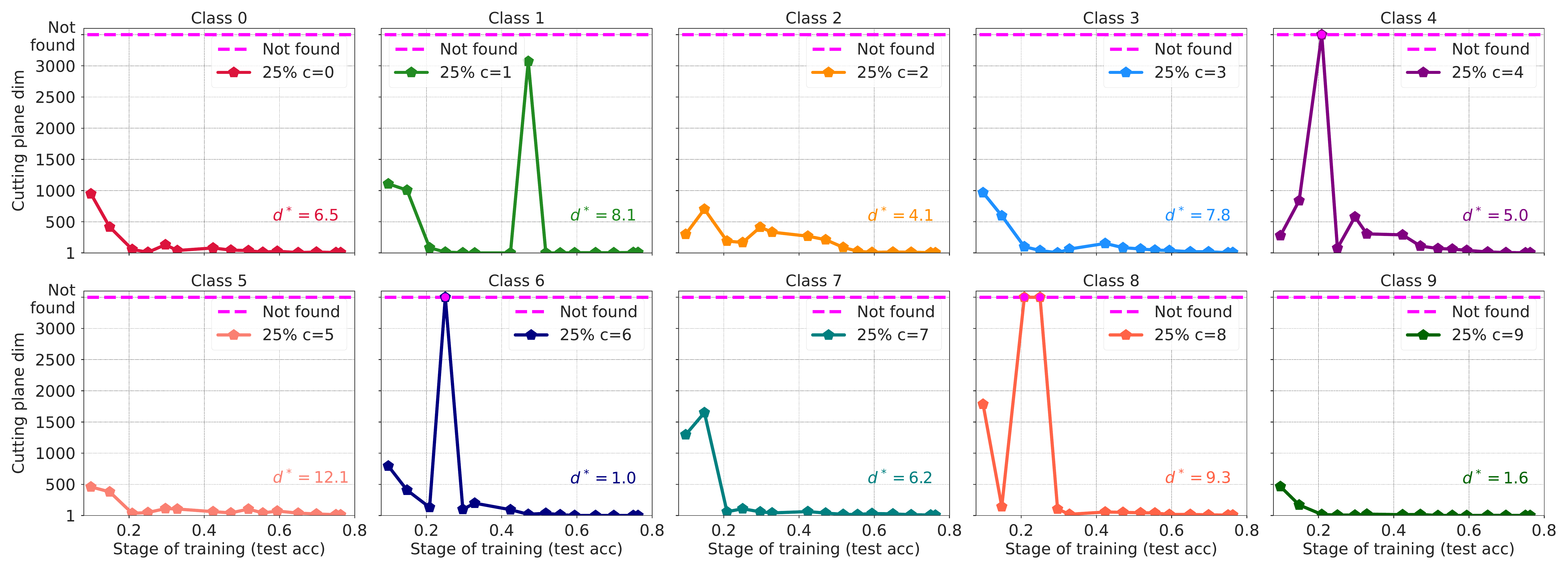}
		\caption{The cutting plane dimension needed to reach $25\%$ probability for the 10 classes of CIFAR-10 as a function of training stage for a SimpleCNN.}
		\label{fig:single_class_dim_vs_test_acc_simplecnn_cifar10}
	\end{figure}

\begin{figure}[ht]
		\centering
		\includegraphics[width=1.0\linewidth]{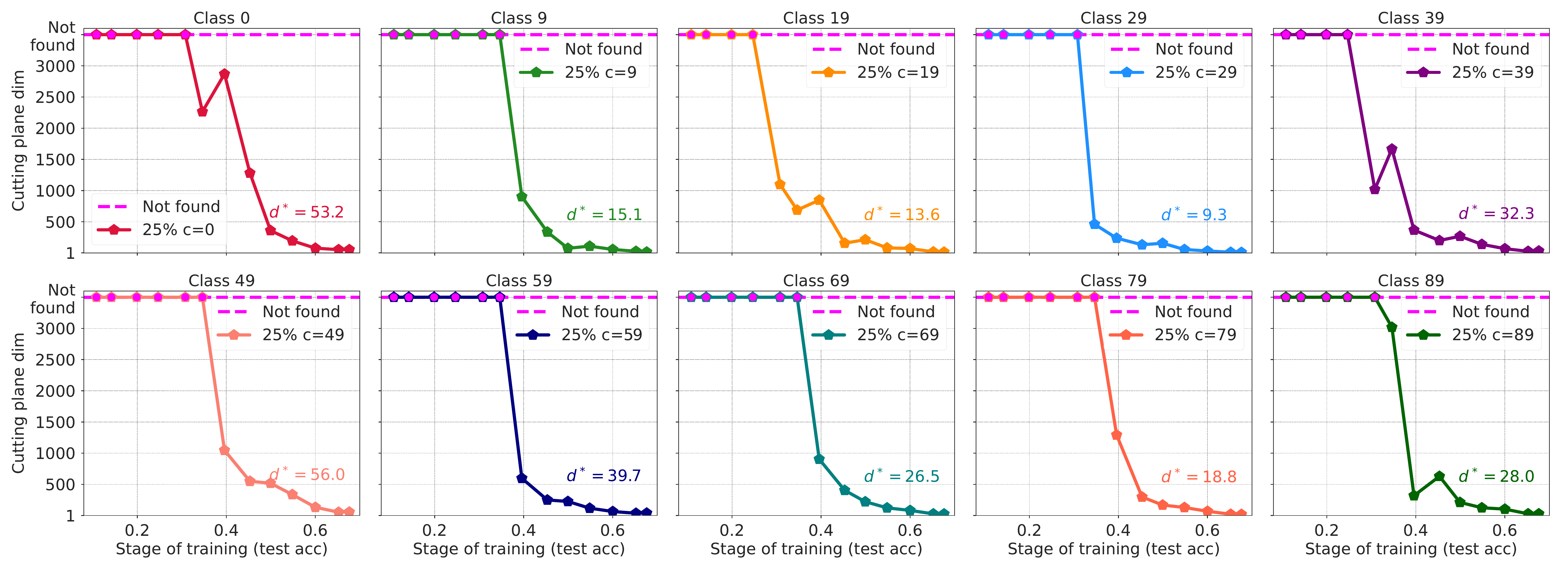}
		\caption{The cutting plane dimension needed to reach $25\%$ probability for 10 randomly selected classes of CIFAR-100 as a function of training stage for a fully trained ResNet20v1.}
		\label{fig:single_class_dim_vs_test_acc_resnet_cifar100}
	\end{figure}

	While Figure \ref{fig:effect_of_training} shows the aggregate effect of training epoch on the the critical cutting plane dimension averaged over all single-class regions, the detailed per-class results can be found in Figure \ref{fig:single_class_dim_vs_test_acc_resnet_cifar10_2inits} for ResNet20v1 on CIFAR-10 (two indepdently initialized and trained models), in Figure \ref{fig:single_class_dim_vs_test_acc_simplecnn_cifar10} for SimpleCNN on CIFAR-10, and in Figure \ref{fig:single_class_dim_vs_test_acc_resnet_cifar100} for ResNet20v1 on CIFAR-100.
	
	\subsection{The effect of network width}
\begin{figure}[ht]
		\centering
		\includegraphics[width=1.0\linewidth]{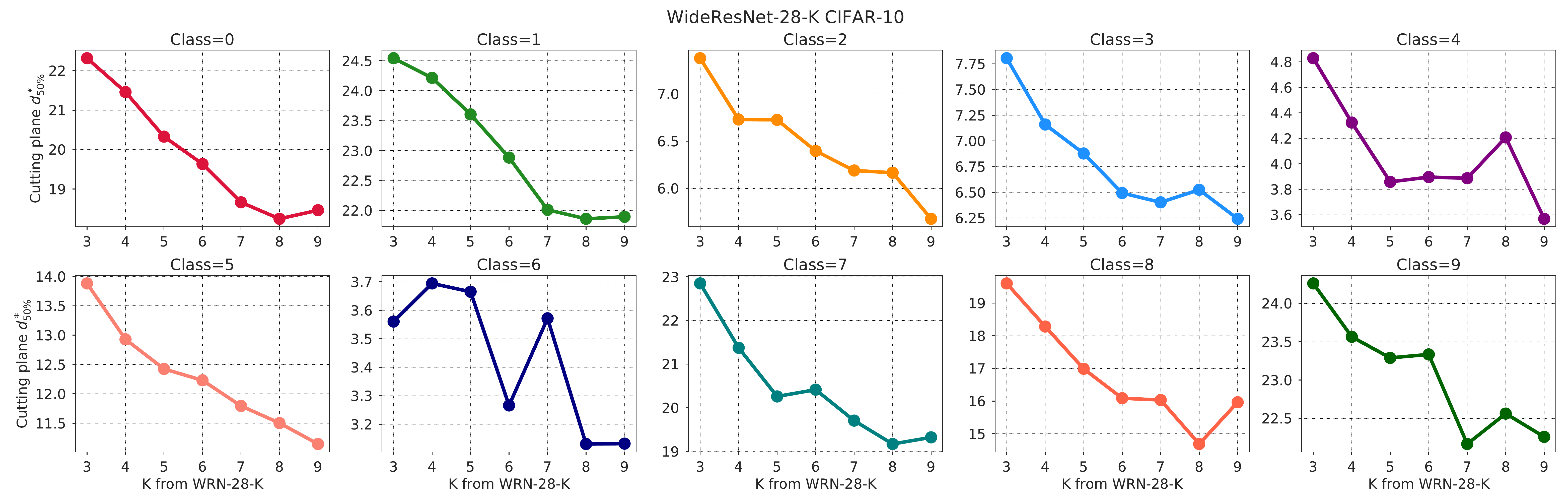}
		\caption{The effect of network width on the dimension of cutting plane necessary to reach a particular probability. The panels shows results for individual classes of CIFAR-10 for WRN-28-K for different values of the width parameter $K$. $d^*_{50\%}$, the dimension of a cutting plane need to reach the class manifolds, goes down with with $K$, meaning that the class manifold dimension goes up as $3072 - d^*_{50\%}$. The accuracy and dimension averaged over all 10 classes are shown in Figure~\ref{fig:effect_of_width_average_class}.
		}
		\label{fig:effect_of_width_all_classes}
	\end{figure}

	We found that wider networks have lower class manifold dimensions. Our results for WideResNet-28-K \citep{zagoruyko2017wide} (WRN-28-K, where $K$ is specifying the width of the layers) averaged over all 10 classes of CIFAR-10 are shown in Figure~\ref{fig:effect_of_width_average_class}. The results for individual classes are shown in Figure~\ref{fig:effect_of_width_all_classes}. The trend that with higher width $K$ the $d^*_{50\%}$ goes down and therefore the manifold dimension goes up holds for individual classes as well as their average.
	
\subsection{The effect of training set size}
\begin{figure}[ht]
		\centering
		\includegraphics[width=1.0\linewidth]{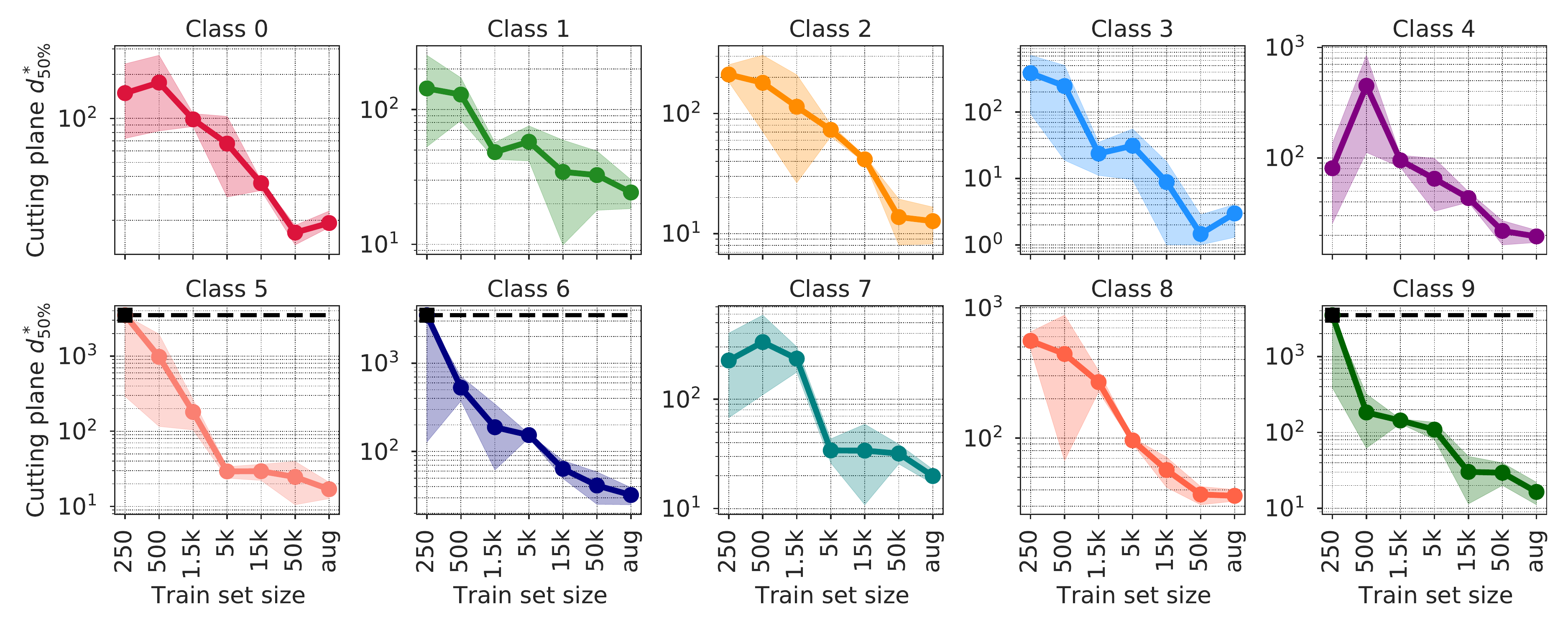}
		\caption{Comparison of the cutting plane dimension needed to get $50\%$ of the target class for ResNets trained to $100\%$ training accuracy on subsets of the training set of CIFAR-10 (mean and standard deviation of 2 networks shown). The bigger the training set, the smaller the $d^*_{\mathrm{50\%}}$, therefore the higher the class manifold dimension. The trend continues with the addition of data augmentation (aug), and takes the manifolds from low-D ($\ll D$) for small sets, to high-D ($\approx D$) for large set + augmentation.}
		\label{fig:train_set_size_effect_all_classes}
	\end{figure}

In Fig.~\ref{fig:train_set_size_effect} we show an example of the training set size dependence of the cutting plane dimension $d^*_{50\%}$ for 3 classes of CIFAR-10. The results for all 10 classes can be found in Fig.~\ref{fig:train_set_size_effect_all_classes}.
	
\subsection{Simple measures of dataset dimensionality}
\label{sec:dataset_dimension}
In this work we focus on measuring the dimension of the learned class manifolds that a trained neural network develops during the course of training on a dataset. Generally, the dimensions we find are very high, for examples look at the summary Figure~\ref{fig:models_plot}. For CIFAR-10 and 100 we observe class dimension manifolds of even 3000 and above out of 3072. To get a comparison between the learned manifold and the dataset itself, we looked at several simple measures of dimension for the dataset itself:
\begin{enumerate}
    \item The number of dimensions in the Principle Components Analysis of the images of a particular class that explain $90\%$ of the variance.
    \item Participation ratio as described in \cite{gao2017theory}
    \item The effective dimension as described in \cite{vershynin_2018} and which we use indirectly to estimate the dimension of the learned manifolds as well.
\end{enumerate}
For the individual classes of CIFAR-10, we get $d_{\mathrm{PCA}} = 98 \pm 20$, $d_{\mathrm{participation}} = 260 \pm 30 $, and $d_{\mathrm{effective}} = 4.5 \pm 0.5 $. All of these estimates are $\ll 3072$ and $\ll$ the measured dimension of the learned class manifolds.

\end{document}